\documentclass{article}

\usepackage[logo]{ace_ego}            
\usepackage{amsmath}
\usepackage{amssymb}
\usepackage{placeins}
\usepackage{graphicx}
\usepackage{xcolor}
\usepackage{multirow}
\usepackage{xspace}
\usepackage{subcaption}
\definecolor{mypurple}{RGB}{160,80,200}
\usepackage{booktabs}
\usepackage[normalem]{ulem}
\usepackage{marvosym}
\definecolor{hsyblue}{RGB}{0,90,200}
\definecolor{hsygray}{RGB}{140,140,140}

\newcommand{\OURS}{\textsc{ACE-Ego-0}\xspace}
\newcommand{\Ours}{\OURS}

\title{\Ours: Unifying Egocentric Human and Robotic Data for VLA Pretraining}

\date{Jun 2026}
\metadata[Website]{\url{https://acerobotics-vla.github.io/ACE-Ego/}}
\metadata[Code]{\url{https://github.com/ACERobotics-VLA/ACE-Ego}}

\author{%
  \sffamily\bfseries\small
  Hao Li$^{1,2*}$ \And
  Ganlong Zhao$^{1,2*,\dagger}$ \And
  Yufei Liu$^{1,4*}$ \And
  Haotian Hou$^{1,2*}$ \And
  Guoquan Ye$^{1,3}$ \And
  Tongyan Fang$^{1,5}$ \AND
  Chunxiao Liu$^{1}$ \And
  Siyuan Huang$^{1\dagger}$ \And
  Jianbo Liu$^{1}$ \And
  Xiaogang Wang$^{1,2}$ \And
  Hongsheng Li$^{2,1{\,\scalebox{0.75}{\normalfont\Letter}}}$
  \\[6pt]
  \mdseries\normalsize
  \begin{tabular}[t]{@{}l@{}}
    $^{1}$ACE Robotics \quad
    $^{2}$CUHK MMLab \quad
    $^{3}$CUHK, Shenzhen \quad
    $^{4}$SJTU \quad
    $^{5}$THU\\[3pt]
    {\footnotesize
    $^*$Equal contribution \quad
    $^\dagger$Project lead \quad
    $^{{\scalebox{0.75}{\normalfont\Letter}}}$Corresponding author}
  \end{tabular}
}

\begin{document}
\maketitle

\begin{abstract}

Vision-Language-Action (VLA) models benefit from large-scale and diverse embodied data, yet scaling robot trajectory collection is costly and labor-intensive.
Recent advances show that large-scale egocentric human videos provide complementary real-world supervision in pretraining.
However, joint training on human and robot data remains challenging due to divergences in action spaces, embodiment structures, temporal dynamics, and supervision quality. 
We introduce \Ours,  a unified VLA pretraining framework jointly leveraging heterogeneous data sources. To extract large-scale pretraining supervision from egocentric human videos, we build a scalable egocentric video-to-action pipeline that converts raw human videos into robot-format pseudo-action trajectories. To make these labels comparable with robot demonstrations, \Ours uses a unified action representation based on camera-space actions, morphology conditioning, and time-aligned action chunking. To robustly leverage noisy pseudo-action supervision from egocentric human videos, we formulate a reliability-aware training objective with a human auxiliary loss that concentrates supervision on reliable signals. 
We instantiate \Ours on 4.53K hours of robot and simulation data, together with 1.48K hours of pseudo-action-labeled egocentric human data. Experiments show that incorporating large-scale human supervision under reliability-aware weighting consistently improves both unified joint pretraining and supervised fine-tuning. 
\Ours achieves state-of-the-art performance on RoboCasa GR1 TableTop and RoboTwin~2.0, while demonstrating strong transfer to real-world bimanual manipulation.

\end{abstract}


\begin{figure}[bhpt] 
    \centering 
\includegraphics[width=\linewidth]{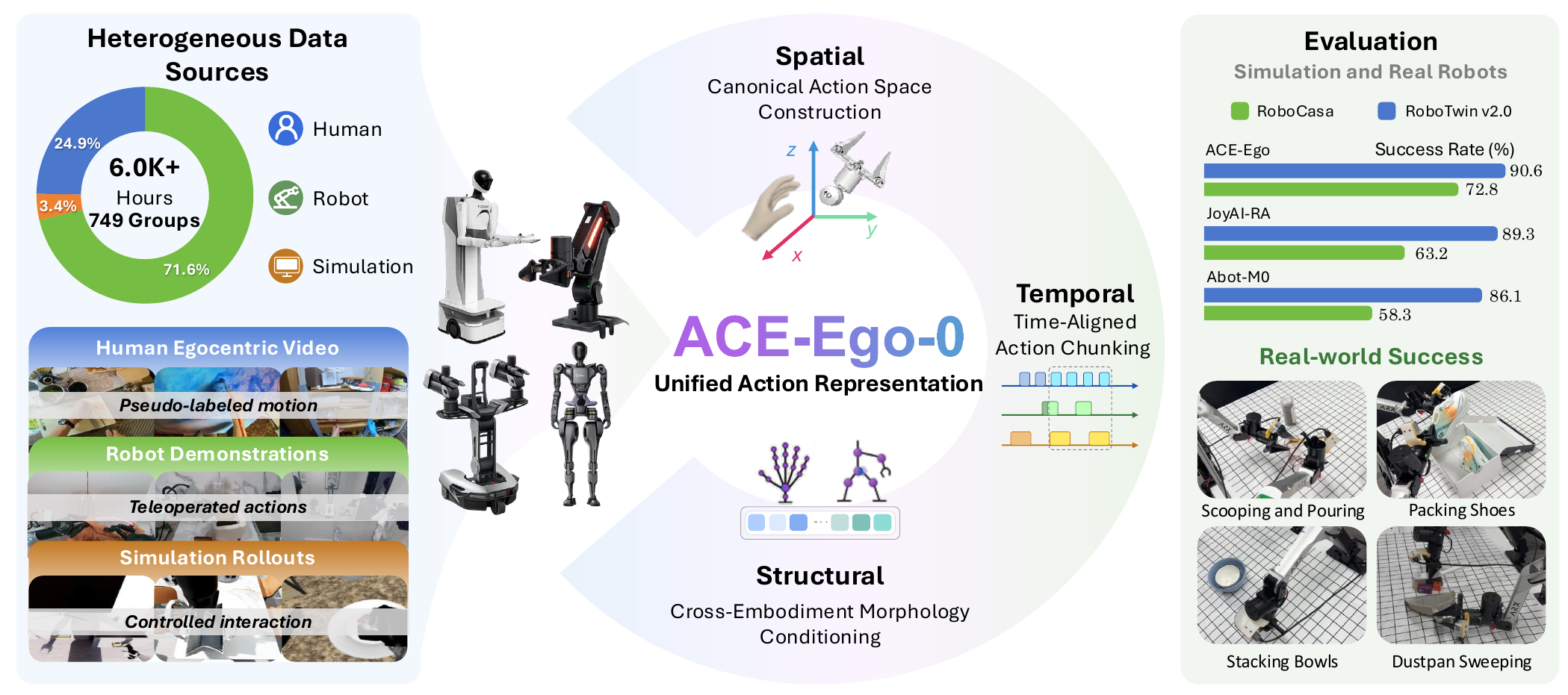} \caption{Overview of \Ours. We pretrain a unified VLA policy on a 6.0K+ hour mixed embodied dataset comprising large-scale egocentric human videos, multi-embodiment robot demonstrations, and simulation rollouts. \Ours unifies heterogeneous human and multi-embodiment robot data into a shared representation space through spatial, structural, and temporal alignment. We achieve state-of-the-art performance on RoboCasa and RoboTwin~2.0, while demonstrating strong real-world bimanual transfer.}
\label{fig:teaser} \end{figure}


\section{Introduction}
\label{sec:introduction}

Developing general-purpose robotic systems capable of operating across diverse real-world environments remains a central objective of embodied AI. Vision-Language-Action (VLA) models~\cite{brohan2023rt,zitkovich2023rt,BlackK-RSS-25,pmlr-v305-black25a} offer a promising path toward this goal by jointly modeling perception, language, and action. A common premise is that broad and diverse embodied experience is critical for acquiring generalizable manipulation skills. Similar to the scaling trends observed in language and vision foundation models, the performance of VLA policies is strongly correlated with the scale and diversity of the training data available during pretraining. However, collecting robot demonstrations at scale remains costly and labor-intensive, limiting both dataset size and behavioral diversity. Large-scale egocentric human videos provide a compelling complementary source of embodied supervision, offering substantially broader coverage of real-world interactions at much lower collection cost. Integrating these heterogeneous data sources into a unified training framework remains challenging due to discrepancies in spatial representations, embodiment structures, temporal horizons, and supervision fidelity.

Existing cross-embodiment VLA methods~\cite{o2024open,octo_2023,wang2024hpt,bjorck2025gr00t,zheng2025x} address representation heterogeneity through shared action spaces, embodiment-specific tokenizers, soft-prompted action experts, or latent action representations, enabling heterogeneous robot demonstrations to be trained within a unified policy framework. However, these approaches remain bottlenecked by the scalability of robot data collection, as they rely primarily on teleoperated demonstrations.
Large-scale egocentric human videos have recently emerged as an appealing complementary source: they are far cheaper to collect and cover a much broader range of manipulation skills in everyday scenes. Several recent works~\cite{bjorck2025gr00t,kareer2024egomimic,yang2025egovla,fu2024humanplus} leverage egocentric human videos, reconstructing hand trajectories and contact targets as action proxies for pseudo-action supervision. However, treating pseudo-actions as equivalent to sensor-logged robotics actions during training injects the label noise directly into the model. In addition, current 3D human hand reconstruction methods typically express hand poses in local space~\cite{pavlakos2024reconstructing,wilor} using MANO~\cite{romero2022embodied}, whereas robot demonstrations are generally recorded in global world space. This misalignment prevents policy models from effectively using both human and robot data for unified policy training. Neither representation heterogeneity nor supervision-quality mismatch is fully resolved in current mixed-source VLA pretraining frameworks.

We present \Ours, a VLA pretraining framework with unified action representation for heterogeneous embodied data, bridging spatial, structural, and temporal discrepancies. Specifically, we introduce canonical action space construction that represents both robot end-effector trajectories and reconstructed human hand pseudo-action trajectories in a common observation-centric coordinate frame, eliminating the need for the policy to learn embodiment-specific coordinate transformations beyond a standard camera extrinsic. To accommodate diverse embodiments, we incorporate cross-embodiment morphology conditioning via embedding robot kinematic descriptions and learned surrogate embeddings for human-video sources. Furthermore, we propose time-aligned action chunking, which indexes future actions according to physical timestamps rather than frame indices, ensuring temporal consistency across datasets collected at different control frequencies. As supervision quality varies substantially across data sources, representation alignment alone is insufficient to achieve effective mixed-source pretraining. We introduce a reliability-aware training objective that explicitly accounts for supervision fidelity. Sensor-logged robot trajectories supervise the primary flow-matching objective, while pseudo-actions are down-weighted and serve as auxiliary supervision primarily on noiseless position channels and modulated by dataset-level and step-level quality estimates. 


We propose a scalable five-stage egocentric data processing pipeline and apply it over six diverse egocentric video datasets to obtain 1.48K hours of pseudo-action-labeled human video. Combining it with 4.53K+ hours of sensor-logged multi-embodiment robot demonstrations and simulation rollouts yields a 6.0K+ hour heterogeneous pretraining dataset for our proposed \Ours. We evaluate \Ours on RoboCasa, RoboTwin~2.0, and a real bimanual ARX platform. \Ours reaches 72.8\% average success on RoboCasa GR1 TableTop benchmark, achieves 91.12\% and 90.62\% average success rates on RoboTwin~2.0 Easy/Hard splits, and demonstrates strong real-world bimanual performance on long-horizon, contact-rich tasks. Ablation studies confirm that morphology conditioning, time-aligned action chunking, and reliability-aware human supervision each contribute to the final performance, and that scaling pseudo-action-labeled human video on top of robot data yields further gains.

Our contributions are summarized as follows.
\begin{itemize}
    \item We introduce \Ours, a unified VLA pretraining framework addressing representation heterogeneity via a unified action representation and supervision-quality mismatch via a reliability-aware training objective.
    \item We develop a scalable five-stage pipeline that converts large-scale egocentric human videos into robot-compatible pseudo-action trajectories, producing 1.48K hours of pseudo-action-labeled human data and enabling joint pretraining with 4.53K hours of multi-embodiment robot and simulation data.
    \item We demonstrate that large-scale human supervision consistently improves both unified VLA pretraining and downstream supervised fine-tuning, achieving state-of-the-art performance on RoboCasa and RoboTwin~2.0 while exhibiting strong transfer to real-world bimanual manipulation.
\end{itemize}


\section{Related Work}

\subsection{Scalable Vision-Language-Action Model Pretraining}
Recent progress in robot learning has moved from task-specific imitation policies toward generalist vision-language-action (VLA) models trained on large and diverse robot datasets.
RT-1~\cite{brohan2023rt} showed that transformer policies can absorb large-scale real-robot demonstrations and generalize across language-conditioned manipulation tasks, and RT-2~\cite{zitkovich2023rt} connected web-scale vision-language pretraining with robot action prediction. 
The Open X-Embodiment and RT-X effort~\cite{o2024open} then aggregated robot trajectories across institutions, embodiments, and task families, establishing cross-embodiment training as a viable route to broader generalization.  
A growing family of open and large-scale VLA systems---including Octo~\cite{octo_2023}, OpenVLA~\cite{kim2025openvla}, $\pi_0$~\cite{BlackK-RSS-25}, $\pi_{0.5}$~\cite{pmlr-v305-black25a}, RDT~\cite{liu2025rdt}, CogACT~\cite{li2024cogact}, and GR00T~\cite{bjorck2025gr00t}---has since scaled model capacity, data diversity, and action-generation flexibility.
However, the very data scaling that fuels these foundation models also introduces a fundamental bottleneck: as a single policy ingests increasingly diverse sources, treating them as a homogeneous corpus becomes exceptionally challenging, because robot datasets differ simultaneously in coordinate frames, kinematic structure, and control frequency.

Prior works have attempted to mitigate this \emph{representation mismatch} along individual axes. Shared end-effector action formats and discrete action tokenizers facilitate cross-dataset training~\cite{o2024open,kim2025openvla}; embodiment-aware tokenizers, adapters, or projectors handle kinematic heterogeneity prior to a shared backbone~\cite{wang2024hpt,bjorck2025gr00t}; and universal or latent action spaces seek to minimize embodiment-specific action discrepancies~\cite{zheng2025uniact,ye2025lapa,liu2026rdt2}. Spatially grounded policies, such as SpatialVLA~\cite{qu2025spatialvla}, 3D-VLA~\cite{zhen2024threedvla}, and TraceVLA~\cite{zheng2024tracevla}, incorporate 3D geometric structures or image-space trajectories to align perception and action. 
Yet, these mechanisms rarely address all three dimensions of heterogeneity jointly: a shared action vector does not guarantee aligned coordinate frames; fixed-length action chunks span disparate physical durations under varying control frequencies; and kinematic structures are often implicitly absorbed via simple dataset IDs or learned codes. 
In contrast, \Ours systematically aligns heterogeneous robot sources across all three axes prior to the shared VLA training objective---employing a unified camera-space action representation, cross-embodiment morphology tokens, and time-aligned action chunking.

\subsection{Learning from Egocentric Human Video}
Beyond robot-collected data, egocentric human video offers a highly scalable and cost-effective source of manipulation experience, capturing rich object interactions, diverse environments, and long-tail behaviors that are difficult to acquire via robot teleoperation. Large-scale egocentric datasets, such as Ego4D~\cite{grauman2022ego4d}, EPIC-KITCHENS~\cite{damen2022epickitchens}, EgoExo4D~\cite{grauman2024egoexo4d}, EgoDex~\cite{hoque2025egodex}, and EgoScale~\cite{zheng2026egoscale}, have significantly amplified this potential. 
Earlier paradigms leveraged such videos primarily for representation or visual reward learning~\cite{nair2022r3m,ma2022vip,ma2023liv,xiao2022mvp,majumdar2023vc1,karamcheti2023voltron,lin2022egovlp,zhao2023lavila}; while these methods extract strong visual priors, they still rely heavily on downstream robot demonstrations to map perception to motor control. 
More recent endeavors extract direct action-level supervision from human videos, either by learning latent or inverse-dynamics actions from action-free footage~\cite{ye2025lapa}, or by reconstructing explicit hand, wrist, or body trajectories and mapping them to robot-compatible commands via retargeting, inverse kinematics, visual domain translation, or morphology-agnostic formulations~\cite{fu2024humanplus,li2024okami,kareer2024egomimic,lepert2025phantom,yang2025egovla,zhu2025emma,li2025h2r,liu2025egozero,bharadhwaj2025zeromimic}.
DIAL~\cite{chen2026dial} takes a different route, incorporating egocentric human video into VLA pretraining through a latent world model that decouples high-level intent prediction from low-level action generation.

Although these advances unlock human video as a scalable supervision source, they expose a critical \emph{supervision-quality mismatch} that is orthogonal to representation heterogeneity. Unlike high-fidelity sensor-logged robot trajectories, human action labels extracted via vision pipelines are inherently noisy pseudo-actions, prone to tracking jitter, occlusions, and estimation bias. 
Existing frameworks typically either bypass direct action-level training or naively feed these noisy pseudo-actions into the same behavior-cloning or diffusion objectives used for clean robot data. This equivalent treatment forces the policy to directly mimic the artifacts and failures of the reconstruction pipeline. 
To resolve this, \Ours routes human-video samples through a reliability-aware auxiliary objective. By restricting supervision to highly reliable position channels and dynamically weighting the loss based on both dataset-level and step-level quality estimates, we ensure that high-fidelity robot data anchor the primary action expert, while human videos provide safe, robust, and complementary auxiliary supervision.

\section{Method}
\begin{figure}[t]
  \centering
  \includegraphics[width=\linewidth]{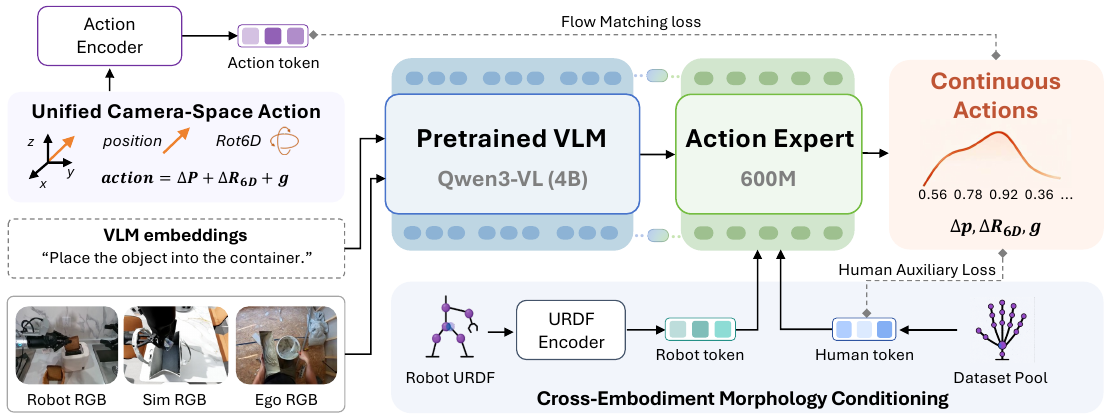}
    \caption{Architecture of \Ours. The vision-language backbone processes multi-view images and language instructions into shared representations. The action expert receives these representations together with morphology tokens that encode each source's embodiment (robot URDF or human surrogate) to predict time-aligned camera-space action chunks via flow matching. Robot samples supervise the primary action loss; human samples contribute through an auxiliary loss with per-channel reliability weighting that concentrates supervision on the position channels.}

  \label{fig:method}
\end{figure}

To pretrain a generalizable VLA policy on heterogeneous embodied data, we must overcome two fundamental challenges: \emph{representation heterogeneity} and \emph{supervision-quality mismatch}. \Ours introduces a two-fold framework as illustrated in Fig.~\ref{fig:method}. First, we establish a \textbf{Unified Action Representation} (Sec.~\ref{sec:unified-interface}) that aligns multi-embodiment data along spatial, structural, and temporal spaces. Second, to prevent estimation noise from human pseudo-actions from corrupting the shared policy in the unified action representation, we propose a \textbf{Reliability-Aware Training Objective} (Sec.~\ref{sec:reliability-aware-objective}) that leverages noisy human pseudo-actions as auxiliary supervision.

\subsection{Unified Action Representation}
\label{sec:unified-interface}

Jointly training on diverse robot trajectories and human videos requires a shared action interface that removes dataset-specific coordinate and temporal conventions. We achieve this by projecting all data sources from three perspectives: \emph{spatial} alignment via camera-space coordinates (Sec.~\ref{sec:camera-space-action}), \emph{structural} alignment via kinematic morphology conditioning (Sec.~\ref{sec:urdf-conditioning}), and \emph{temporal} alignment via time-aligned action chunking (Sec.~\ref{sec:time-aligned-dynamic-chunking}). Together, these mechanisms map heterogeneous trajectories into a single, embodiment-agnostic action space.

\subsubsection{Canonical Action Space}
\label{sec:camera-space-action}

\Ours first aligns sources spatially by representing actions from both robot and human data in the head-camera coordinate frame before they enter the model. Predicting in the camera frame keeps actions and observations in a unified coordinate system, eliminating the need for the policy to learn complex, platform-specific world-to-camera transformations. Under this formulation, actions and observations are fed in a platform-agnostic framework, and the pretrained policy transfers to a new embodiment by simply swapping a single camera extrinsic at inference time.

\paragraph{Robot action convention.}
For each robot source, the bimanual end-effector poses are projected into the head-camera frame. Poses on top of a robot base or in a world frame $s$ are transformed using the calibrated camera extrinsic:

\begin{equation}
  p_{\mathrm{cam}} = R_{\mathrm{cam}\leftarrow s}\, p_s + t_{\mathrm{cam}\leftarrow s},
  \qquad
  R_{\mathrm{cam},ee} = R_{\mathrm{cam}\leftarrow s}\, R_{s,ee},
  \label{eq:cam-transform}
\end{equation}
where $p_s$ and $R_{s,ee}$ denote the end-effector position and orientation in the source frame, respectively. Orientations are parameterized using a continuous 6D representation~\cite{zhou2019continuity}. Combined with gripper commands and arm activity flags, this yields a unified bimanual action vector (see Appendix~\ref{app:robot-action-standardization} for the details). Expressing actions in this unified format ensures that the action expert consumes both human and robot trajectories through a shared states interface.

\paragraph{Human end-effector equivalents.}
Since human hands do not have a physical end-effector, we define a hand-centric coordinate frame as proxy end-effector, allowing human motions to be represented in a robot-compatible form while remaining directly connected to hand mesh reconstruction.
We designate the wrist joint as the end-effector origin, as it is reconstructed most consistently in HaMeR's~\cite{pavlakos2024reconstructing} frame-wise predictions. To mitigate yaw drift under occlusion, we construct a stable hand-centric orientation frame $R \in \mathrm{SO}(3)$ using the palm plane and wrist-to-finger vectors, which is then converted to the same continuous 6D representation used for robots. For gripper openness, we employ the normalized thumb-to-palm distance as a proxy for hand closure, linearly scaled to match the robot's physical gripper stroke. This parameterization normalizes human trajectories using the base value and maps them into the shared bimanual action space, which is used for seamless joint training across human and robot data. The exact geometric derivations of the hand frame are detailed in Appendix~\ref{app:robot-action-standardization}.

\subsubsection{Cross-Embodiment Morphology Conditioning}
\label{sec:urdf-conditioning}

Although a canonical action space resolves spatial discrepancies, differences in kinematic chains, joint limits, and physical dimensions still persist across embodiments. To unify the cross-embodiment discrepancy, we embed humans and each robot type into a shared morphology space. A major contribution of \Ours is addressing this structural mismatch by conditioning the action expert on a morphology token. For robots, this token is dynamically computed from its URDF graph; for humans, it is shared across different people and updated via back-propagation. Crucially, we keep this morphology token isolated from the vision-language backbone and inject it only during action decoding, thereby keeping our VLM backbone embodiment-agnostic.
Robot and human morphologies are projected into this shared token space via parallel pathways:

\begin{equation}
  h_{\mathrm{morph}}
  =
  \begin{cases}
  P_{\mathrm{morph}}\!\left(E_{\mathrm{urdf}}(\mathcal{G}_{r})\right),
    & \text{robot source } r, \\[4pt]
  P_{\mathrm{surr}}(e_{d}),
    & \text{human source } d,
  \end{cases}
  \label{eq:morph-token}
\end{equation}
where $E_{\mathrm{urdf}}$ encodes the URDF graph $\mathcal{G}_r$ at both global and local manipulation scales (Appendix~\ref{app:urdf-graph-construction}), and $e_d$ is a learned surrogate embedding capturing the visual and dataset-specific priors of human source $d$ (Appendix~\ref{app:human-surrogate-details}). Both pathways condition the action expert through a unified interface.


\subsubsection{Time-Aligned Action Chunking}
\label{sec:time-aligned-dynamic-chunking}

For temporal alignment, robot datasets often have different control frequencies.
If we predict a fixed number of future steps, the policy must plan for different physical durations across datasets. To prevent this temporal mismatch, \Ours defines action chunks by physical duration rather than step count.
For a dataset \(d\) with control frequency \(f_d\), we set the step horizon \(H_d\) based on a target physical duration \(T^\star\):
\begin{equation}
H_d = \mathrm{round}\!\left(f_d T^\star\right).
\label{eq:horizon}
\end{equation}
This formulation ensures that all datasets supervise the same future physical window \(T^\star\). However, training on variable-horizon chunks within the same batch can cause large padding overhead and training instability. We address these issues with a structured batch sampling strategy. Specifically, trajectories are pre-chunked according to the target physical window to maintain temporal consistency and minimize padding overhead. For a sample starting at index \(t\) in an episode of length \(L_e\), we define the normalized episode phase \(\phi\) as:
\begin{equation}
\phi = \operatorname{clip}\!\left( \frac{t + \tfrac{1}{2}H_d}{L_e}, 0, 1 \right),
\label{eq:episode-phase}
\end{equation}
Since \(H_d\) is determined by the target physical duration and dataset control frequency, \(\phi\) is comparable across datasets with different frame rates. We discretize \(\phi\) into a phase bucket \(b_{\phi}\) and \(H_d\) into a horizon bucket \(b_H\). We then form mini-batches using a composite key:
\begin{equation}
k = \left(c_{\mathrm{task}}, b_{\phi}, b_H\right),
\label{eq:batch-key}
\end{equation}
where \(c_{\mathrm{task}}\) is a task cluster from episode metadata. 
This bucketing strategy balances training stability and computational efficiency. Grouping by task ensures semantic coherence within each batch, while grouping by horizon minimizes the padding required for samples with different chunk lengths, thereby significantly reducing padding overhead and stabilizing the gradient updates.



\subsection{Reliability-Aware Training Objective}
\label{sec:reliability-aware-objective}

Even with aligned action spaces, naive joint training on mixed-source data risks propagating estimation noise from human pseudo-actions directly into the action expert, which degrades the learning of the robust control policy from high-fidelity robot data. To resolve this supervision-quality mismatch, we propose a reliability-aware training objective. 
We formally define the spatiotemporal reliability for each action dimension (e.g., control channel) \(j \in \{1, \dots, D\}\) at step \(t\) as:
\begin{equation}
W_{t,j} = \rho_j \cdot w_{t,j},
\label{eq:reliability}
\end{equation}
where \(\rho_j \in [0, 1]\) is a static, channel-level prior reflecting the intrinsic tracking stability of different action dimensions. In practice, these priors \(\rho_j\) are empirically assigned based on the measurement noise of the human pose estimator (e.g., positioning channels are highly reliable and assigned \(\rho=1.0\), whereas wrist rotations and gripper states are prone to occlusion noise and assigned lower weights). The term \(w_{t,j} \in [0, 1]\) represents a dynamic, step-level smoothness factor that down-weights local tracking failures or implausible kinematic jumps. 

With this reliability-aware weighting strategy, high-fidelity robot data anchors the primary objective across all channels, while noisy human pseudo-actions contribute to training through a robust auxiliary loss scaled by \(W_{t,j}\). The dynamic term \(w_{t,j}\) further factorizes into a dataset-level prior and a local step-level smoothness weight, with exact formulations detailed in Appendix~\ref{app:reliability-aware-human-details}.

\paragraph{Robot Primary Loss}

The primary robot loss follows the standard conditional flow-matching formulation, optimized over the valid action dimensions selected by the action mask $M$. Given a clean, sensor-logged robot action target $\mathbf{a}$ and Gaussian noise $\boldsymbol{\epsilon} \sim \mathcal{N}(0, I)$, the flow interpolant is defined as $\mathbf{a}_s = s\mathbf{a} + (1{-}s)\boldsymbol{\epsilon}$ for $s \sim \mathcal{U}(0,1)$. Then the robot loss is formulated as:

\begin{equation}
\label{equ:action_loss}
\mathcal{L}_{\mathrm{action}}
=
\mathbb{E}_{s,\boldsymbol{\epsilon}}
\sum_{t,j}
M_{t,j}
\left\|
\hat{v}_\theta(\mathbf{a}_s, s)_{t,j}
-
(\mathbf{a} - \boldsymbol{\epsilon})_{t,j}
\right\|^2,
\end{equation}
where $\hat{v}_\theta$ is the predicted velocity field and $M_{t,j} \in \{0,1\}$ is the action mask. During training, we use the delta action chunk formulation following~\cite{BlackK-RSS-25}, expressed in the head-camera frame.

\paragraph{Human Auxiliary Loss}
\label{sec:human-auxiliary-loss}

To incorporate human demonstrations without corrupting the policy's primary control capabilities, we introduce human auxiliary loss. Let $\tilde{\mathbf{a}}$ denote the temporally smoothed human target, and $\mathbf{a}_s = s\tilde{\mathbf{a}} + (1{-}s)\boldsymbol{\epsilon}$ be the corresponding flow interpolant. We apply the spatiotemporal reliability weight $W_{t,j}$ within a robust Huber regression loss:

\begin{equation}
  \mathcal{L}_{\mathrm{haux}}
  =
  \mathbb{E}_{s,\boldsymbol{\epsilon}}\,
  \frac{1}{Z}
  \sum_{t,j}
  M_{t,j}\,
  W_{t,j}\,
  \operatorname{Huber}_{\beta}\!\left(
  \hat{v}_\theta(\mathbf{a}_s, s)_{t,j}
  -
  (\tilde{\mathbf{a}} - \boldsymbol{\epsilon})_{t,j}
  \right),
\end{equation}
where $Z = \sum_{t,j} M_{t,j} W_{t,j}$ is the normalization factor. This formulation concentrates human supervision on highly reliable position channels while safely discounting noisy rotation and gripper signals (see Appendix~\ref{app:reliability-aware-human-details} for details on the smoothness statistics and thresholds).

The joint training objective is a weighted combination of the two losses:
\begin{equation}
\mathcal{L} = \mathcal{L}_{\mathrm{action}} + \lambda_{\mathrm{haux}}\, \mathcal{L}_{\mathrm{haux}},
\end{equation}
where $\lambda_{\mathrm{haux}}$ balances the contribution of the human auxiliary loss (hyperparameters and sensitivity analyses are provided in Appendix~\ref{app:reliability-aware-human-details}).

\section{Heterogeneous Pretraining Data}
\label{sec:data}

\begin{figure}[t]
  \centering
  \includegraphics[width=\linewidth]{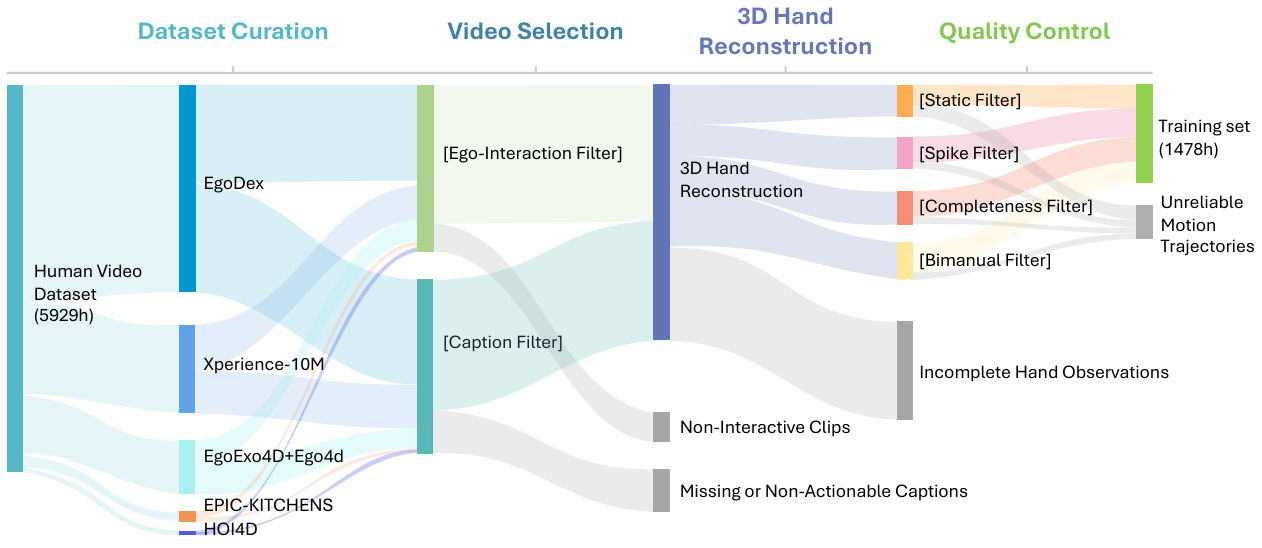}
  \caption{Overview of the \Ours data processing pipeline for constructing training-ready embodied manipulation data from large-scale egocentric human video. Raw videos pass through video selection, motion reconstruction, and multi-stage quality control, yielding 1,478 hours of pseudo-action-labeled embodied manipulation data that complement the robot and simulation portions of the training pool.}
  \label{fig:statics}
\end{figure}

The \Ours pretraining pool covers the full spectrum of embodied experience, including sensor-logged robot demonstrations across multiple platforms, simulation rollouts, and pseudo-action-labeled egocentric human videos, totaling more than 6.0K hours as shown in Table~\ref{tab:pretraining-data-main}. These sources exhibit significant representation heterogeneity as identified in Section~\ref{sec:introduction}: they differ in spatial coordinate frames, kinematic structures, and control frequencies, and further vary in action-label quality.
Section~\ref{sec:data-sources} catalogs our overall mixed-source datasets, while Section~\ref{sec:human-pipeline} describes the pipeline that converts raw egocentric videos into pseudo-action labels compatible with the unified interface defined in Section~\ref{sec:unified-interface}. Figure~\ref{fig:statics} provides a visual overview of this conversion process.

\subsection{Heterogeneous Data Sources}
\label{sec:data-sources}

\paragraph{Robot demonstrations and simulation.}
The robot portion consists of AgiBot Alpha/Beta demonstrations, Galaxea R1Lite data, AgiBot DigitalWorld simulation rollouts, RoboCasa Tabletop simulation data (24 tasks, 1,000 episodes each, GR1 humanoid robot), and 1,800+ hours of self-collected Galbot demonstrations. These platforms span humanoid (AgiBot G1), single-arm wheeled (Galaxea R1Lite), and mobile bimanual (Galbot) embodiments, with control frequencies ranging from 10 to 30~Hz and end-effector poses logged in different reference frames depending on the platforms. This heterogeneity exposes representation mismatch and motivates the unified interface introduced in Section~\ref{sec:unified-interface}. All sources provide sensor-grounded end-effector action labels, which serve as high-fidelity supervision for the primary action expert in Section~\ref{sec:reliability-aware-objective}.

\paragraph{Human egocentric videos.}
The human-video portion draws from six sources: Ego4D~\cite{grauman2022ego4d}, EgoExo4D~\cite{grauman2024egoexo4d}, EPIC-KITCHENS-100~\cite{damen2018epic}, HOI4D~\cite{liu2022hoi4d}, EgoDex~\cite{hoque2025egodex}, and Xperience-10M~\cite{xperience_10m}. Together, they span diverse kitchens, homes, and workshops, capturing long-tail manipulation behaviors that are difficult to cover via robot teleoperation alone. Since their action labels are inferred from vision-based pipelines rather than physical sensors, we treat them as pseudo-action-labeled supervision and route them through the reliability-aware human objective in Section~\ref{sec:human-auxiliary-loss}.

\begin{table}[h]
\centering
\small
\caption{\Ours pretraining data pool. Hours are computed from dataset metadata as $\text{frames} / (\text{fps} \times 3600)$; Galbot hours are reported from our self-collected collections.}
\label{tab:pretraining-data-main}
\begin{tabular}{lrrrr}
\toprule
Source & Episodes & Frames & Hours & Supervision \\
\midrule
Ego4D & 948,683 & 23,396,157 & 216.6 & Pseudo-action \\
EgoExo4D & 41,414 & 1,110,275 & 10.3 & Pseudo-action \\
EPIC-KITCHENS-100 & 74,788 & 3,486,432 & 32.3 & Pseudo-action \\
HOI4D & 2,966 & 774,275 & 7.2 & Pseudo-action \\
EgoDex & 327,317 & 83,894,075 & 776.8 & Pseudo-action \\
Xperience-10M & 99,027 & 31,370,900 & 435.7 & Pseudo-action \\
\midrule
Human video subtotal & 1,494,195 & 144,032,114 & 1,478.9 & Pseudo-action \\
\midrule
AgiBot Alpha/Beta & 116,013 & 209,284,239 & 1,937.8 & Robot action \\
Galaxea R1Lite & 20,662 & 26,358,560 & 488.1 & Robot action \\
AgiBot DigitalWorld & 47,910 & 24,333,788 & 225.3 & Robot action \\
RoboCasa Tabletop & 24,000 & 6,020,058 & 83.6 & Robot action \\
Galbot self-collected & $\sim$60,000 & $\sim$194M & 1,800+ & Robot action \\
\midrule
Robot subtotal & $\sim$268,585 & $\sim$460M & 4,534.8+ & Robot action \\
\midrule
Total & $\sim$1,762,780 & $\sim$604M & 6,013.7+ & Mixed \\
\bottomrule
\end{tabular}
\end{table}

\subsection{Egocentric Video-to-Action Conversion}
\label{sec:human-pipeline}

\begin{figure}[htbp]
  \centering
  \includegraphics[width=\linewidth]{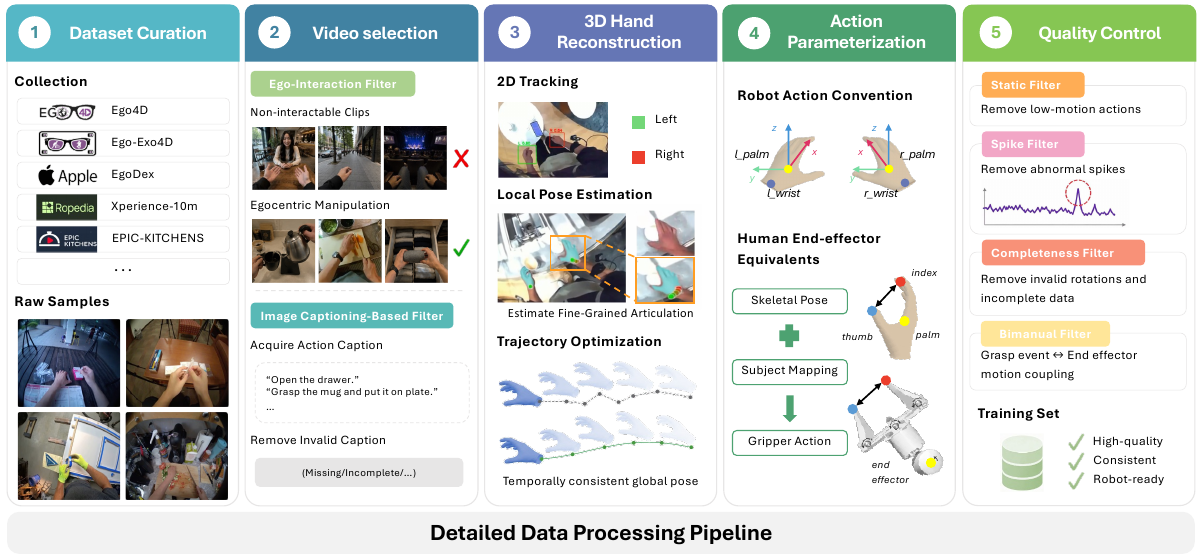}
  \caption{Pipeline for converting raw egocentric videos into camera-space pseudo-actions. The pipeline consists of five stages: (1)~dataset curation;
  (2)~video selection using ego-interaction and image captioning-based filters;
  (3)~3D hand reconstruction, including 2D tracking, local pose estimation, and trajectory optimization;
  (4)~action parameterization using robot action conventions and human end-effector equivalents; and
  (5)~quality control through multiple filtering.}
  \label{fig:data-pipeline}
\end{figure}

Generating pseudo-labeled actions compatible with robotic data from large-scale video datasets requires bridging two major challenges: the \textit{structural discrepancy}, since 2D video carries no metric 3D hand trajectories, and the \textit{behavioral discrepancy}, since not every clip contains a clean manipulation primitive worth supervising on. We address both with a five-stage pipeline (Figure~\ref{fig:data-pipeline})
that includes clip-level filtering, geometric recovery, action formatting, and fidelity-based quality control. Running this pipeline over six egocentric video datasets, we produce $1{,}478$ hours of pseudo-action-labeled clips that share the same camera-space action format as robot data and enter the unified action space of Section~\ref{sec:unified-interface}. All quantitative
thresholds are collected in Table~\ref{tab:egocentric-pipeline-hparams}. We describe each stage in detail below.

\paragraph{Stage~1: Dataset curation.}
We begin with publicly available human video collections and select sources that satisfy three criteria: an egocentric viewpoint, diverse real-world interaction scenes, and high-quality action-centric captions. This process yields the six datasets listed in
Table~\ref{tab:pretraining-data-main}, which forms our human video pool. We then standardize all sources into a unified storage format with consistent metadata fields, including clip identifiers, frame indices, camera intrinsics (when available), narrations, and licensing
information. For sources that provide only video-level annotations, we split videos into clips. We discard clips that are shorter than 4 seconds or longer than 30 seconds, as they are unlikely to contain complete manipulation primitives at the downstream temporal granularity.

\paragraph{Stage~2: Video selection.}
The previous stage produces a large pool of egocentric videos that vary substantially in interaction quality and manipulation relevance. Before applying computationally intensive geometric reconstruction, we adopt an
\emph{ego-interaction filter} to remove clips that are unlikely to provide useful action supervision. The filter targets videos with limited human-object interaction and employs several lightweight cues to identify such cases. Among them, strong face detections serve as an effective signal of non-egocentric or observer-centric viewpoints, which rarely contain usable manipulation trajectories. We therefore discard clips whose maximum face-detection confidence exceeds a predefined threshold. 
A subsequent \emph{image captioning-based filter} retains only clips whose narrations contain at least one manipulation verb and one manipulable object noun, further enriching the dataset with object-centric interaction behaviors.

\paragraph{Stage~3: 3D hand reconstruction.}
Hand reconstruction is performed in three sub-stages: 2D tracking, local pose estimation, and global trajectory optimization. We first apply a SAM3-based~\cite{sam3} tracker to obtain temporally consistent hand bounding boxes and segmentation masks throughout each clip, and discard detections with keypoint confidence below $\tau_{\mathrm{kp}}$ or track length below $\ell_{\min}$ frames. We then feed the retained hand crops into HaMeR~\cite{pavlakos2024reconstructing}, which reconstructs MANO shape and pose parameters $\{\beta, \theta_t, \mathbf{t}^{\mathrm{local}}_t\}_{t=1}^{T}$ for each frame in hand-related clips. Since per-frame reconstruction suffers from depth ambiguity, occlusions, and temporal jitter, we further perform a two-stage global trajectory optimization inspired by~\cite{dyn}, where the first stage ($N_{\mathrm{root}}$ iterations) estimates globally consistent root translation and orientation, and the second stage ($N_{\mathrm{smooth}}$ L-BFGS iterations) jointly minimizes reprojection error and a temporal smoothness regularizer:
\begin{equation}
\mathcal{L}_{\mathrm{smooth}}
=
\mathcal{L}_{\mathrm{reproj}}
+
\lambda_{\mathrm{tv}}
\sum_t
\left\|
\mathbf{t}^{\mathrm{global}}_{t+1}
-
2\mathbf{t}^{\mathrm{global}}_{t}
+
\mathbf{t}^{\mathrm{global}}_{t-1}
\right\|_2^2 ,
\end{equation}
where $\mathcal{L}_{\mathrm{reproj}}$ denotes the 2D keypoint reprojection loss, $\mathbf{t}^{\mathrm{global}}_t$ is the optimized global hand root translation at frame $t$, and $\lambda_{\mathrm{tv}}$ controls the strength of temporal smoothness regularization. Both optimization stages leverage per-frame camera poses $(\mathbf{R}_t^{\mathrm{cam}}, \mathbf{t}_t^{\mathrm{cam}})$ estimated by VIPE~\cite{vipe}, enabling conversion of local reconstructions into temporally coherent 3D trajectories in a shared world coordinate frame. The optimized global trajectory is used only for temporal consistency; the final pseudo-action labels are transformed back into the corresponding head-camera frame before training.

\paragraph{Stage~4: Action parameterization.}
The parameterization itself (wrist origin, palm-plane orientation,
thumb-to-palm gripper proxy) is defined in
Section~\ref{sec:camera-space-action}. Here we explain two implementation
details. \emph{Storage layout.} On disk, each per-hand action is stored as
a 16-dimensional bimanual vector: $3$ position $+ 3$ XYZ Euler $+ 1$
gripper $+ 1$ activity flag, per hand; $8\text{D} \times 2\text{ hands} = 16\text{D}$ total. At training time the Euler angles
are converted to the continuous 6D rotation
representation~\cite{zhou2019continuity}, producing the $22$-dimensional
action vector, defined in Section~\ref{sec:camera-space-action}.
\emph{Gripper normalization.} Thumb-to-palm distances $d_t$ are linearly
normalized to the robot gripper stroke range: 
$[d_{\min}^{\mathrm{grip}}, d_{\max}^{\mathrm{grip}}]
= [0.04, 0.10]\,\mathrm{m}$. Trajectories whose 10th--90th percentile range
satisfies $d_{90} - d_{10} < \tau_{\mathrm{grip}}$ are treated as
degenerate (e.g., closed-fist motion with no grasp transition) and assigned
a constant neutral gripper state.

\paragraph{Stage~5: Quality control.}
This stage removes corrupted or behaviorally implausible human episodes before they are collected into the mixed-source pretraining datasets. Here we apply four post-processing filters. 
\emph{Completeness filter.} We require each
episode to be free of NaN/Inf values, contain contiguous frame indices, and satisfy quaternion normalization constraints: $|\|q\|-1| \le \tau_{\mathrm{quat}}$. \emph{Static filter.} We discard episodes when neither hand exhibits per-second motion energy above $\tau_{\mathrm{static}}$, indicating little or no meaningful interaction. 
\emph{Spike filter.} We reject trajectories if inter-frame positional changes exceed $\kappa_{\mathrm{spike}}\sigma$ of the per-episode velocity distribution on more than $\rho_{\mathrm{spike}}$ of frames, which typically indicates tracking failures or reconstruction artifacts.
\emph{Bimanual filter.} We remove episodes with implausible dual-arm behaviors based on anomalous inter-hand distance statistics or weak temporal correlation between the two hands. We record the corresponding thresholds in the released data manifests since they depend on source-level hand-detection density.

\begin{table}[!t]
\centering
\small
\caption{Egocentric video pipeline hyperparameters used in Stages~1--5. Values are shared across the six human-video sources
unless noted otherwise.}
\label{tab:egocentric-pipeline-hparams}
\renewcommand{\arraystretch}{1.1}
\begin{tabular}{lll}
\toprule
\textbf{Stage} & \textbf{Hyperparameter} & \textbf{Value} \\
\midrule
\multirow[t]{2}{*}{Stage 1: Curation}
  & Min clip duration                               & 4~s \\
  & Max clip duration                               & 30~s \\
\midrule
\multirow[t]{2}{*}{Stage 2: Selection}
  & Face-detection threshold                        & 0.5 \\
  & Caption verb/noun requirement                   & both present \\
\midrule
\multirow[t]{5}{*}{Stage 3: Reconstruction}
  & Keypoint confidence $\tau_{\mathrm{kp}}$        & 0.4 \\
  & Min track length $\ell_{\min}$                  & 15 frames \\
  & Root-fitting iterations $N_{\mathrm{root}}$     & 30 \\
  & Smooth-fitting iterations $N_{\mathrm{smooth}}$ & 200 \\
  & Smoothness weight $\lambda_{\mathrm{tv}}$       & 1.0 \\
\midrule
\multirow[t]{3}{*}{Stage 4: Parameterization}
  & Gripper stroke range $[d_{\min}^{\mathrm{grip}}, d_{\max}^{\mathrm{grip}}]$ & $[0.04, 0.10]\,\mathrm{m}$ \\
  & Gripper-degeneracy threshold $\tau_{\mathrm{grip}}$ & 1.5~cm \\
  & On-disk action dim / training action dim        & 16-D / 22-D \\
\midrule
\multirow[t]{4}{*}{Stage 5: Filtering}
  & Quaternion tolerance $\tau_{\mathrm{quat}}$     & $10^{-3}$ \\
  & Static motion energy $\tau_{\mathrm{static}}$   & source-specific \\
  & Spike $\sigma$-multiplier $\kappa_{\mathrm{spike}}$ & 3 \\
  & Spike frame fraction $\rho_{\mathrm{spike}}$    & 5\% \\
\bottomrule
\end{tabular}
\end{table}


\section{Experiments}

\subsection{Experimental Setup}
\label{sec:exp_setup}
We evaluate \Ours on two simulation benchmarks and one real-robot platform: RoboCasa GR1 TableTop~\cite{bjorck2025gr00t}, a humanoid tabletop benchmark with 24 pick-and-place and articulated-object tasks; RoboTwin~2.0~\cite{robotwin2.0}, a bimanual benchmark with 50 tasks and strong domain randomization; and an ARX bimanual platform with six real-world manipulation tasks. For simulation evaluation, we compare against GR00T-N1.6~\cite{bjorck2025gr00t}, Qwen3PI, FLARE~\cite{pmlr-v305-zheng25a}, ABot-M0~\cite{yang2026abot}, JoyAI-RA~\cite{zhang2026joyaira}, and DIAL~\cite{chen2026dial} on RoboCasa, as well as $\pi_{0.5}$~\cite{pmlr-v305-black25a}, Motus~\cite{bi2025motusunifiedlatentaction}, LingBot-VLA~\cite{wu2026pragmatic}, ABot-M0~\cite{yang2026abot}, JoyAI-RA~\cite{zhang2026joyaira}, and Hy-VLA~\cite{hunyuan2026hyembod} on RoboTwin~2.0 (full per-task comparisons including $\pi_0$ are in Appendix~\ref{app:full-robotwin-results}). For the physical real-robot evaluation, we compare against fine-tuned $\pi_{0.5}$ and GR00T-N1.7~\cite{bjorck2025gr00t}, adopting the N1.7 version to leverage its latest optimizations for physical deployment. All models are trained in a multi-task setting and evaluated by task success rate. Model architecture, training protocol, and evaluation details are provided in Appendix~\ref{app:training-details}.

\paragraph{Camera-space inference.}
To execute these camera-space action chunks on a physical robot, we apply the inverse of the camera extrinsic used during data standardization:
\begin{equation}
\hat{p}_s = R_{\mathrm{cam}\leftarrow s}^{\top}\!\left(\hat{p}_{\mathrm{cam}} - t_{\mathrm{cam}\leftarrow s}\right),
\qquad
\hat{R}_{s,ee} = R_{\mathrm{cam}\leftarrow s}^{\top}\hat{R}_{\mathrm{cam},ee},
\end{equation}
where $s$ denotes the robot's execution frame (e.g., base or torso). The 6D rotation output is first reconstructed into a full rotation matrix via Gram--Schmidt orthogonalization before applying the inverse transform.
Because \Ours predicts actions in the head-camera coordinate frame, deployment only requires a standard extrinsic transform to convert predicted camera-frame end-effector poses into the robot coordinate frame. A new embodiment can be integrated by registering its URDF to obtain a morphology token and executing the resulting poses with its own low-level controller. Thus, camera-space prediction removes the need for the policy to learn source-specific coordinate transforms, while embodiment differences are handled by morphology conditioning.

\subsection{Simulation Results}
\label{sec:sim_results}

\subsubsection{RoboCasa GR1 TableTop}
\label{sec:robocasa}

RoboCasa GR1 TableTop evaluates humanoid tabletop manipulation on the GR1 platform across 24 tasks: 18 pick-and-place rearrangement tasks and 6 articulated-object interaction tasks. 
We train one model jointly on all 24 tasks and report mean success rate over 50 rollouts per task.

\begin{table*}[t]
\centering
\caption{Evaluation results on the RoboCasa GR1 TableTop benchmark (selected tasks). Success rates (\%) over 50 rollouts per task. Full 24-task results are in Appendix~\ref{app:full-robocasa-results}.}
\label{tab:robocasa_gr1_results}
\vspace{0.3em}
\small
\renewcommand{\arraystretch}{1.1}
\setlength{\tabcolsep}{4pt}
\begin{tabular*}{\textwidth}{@{\extracolsep{\fill}}lccccccc}
\hline
\textbf{Task} & GR00T-N1.6 & Qwen3PI & FLARE & ABot-M0 & JoyAI-RA & DIAL & \Ours \\
\hline
CuttingboardToCardboardbox & 46.5 & 46.0 & 54.0 & 58.0 & 46.0 & -- & \textbf{84.0} \\
PlacematToTieredshelf      & 28.5 & 28.0 & 26.0 & 26.0 & 14.0 & -- & \textbf{44.0} \\
PlateToPlate               & 78.7 & 48.0 & 76.0 & 64.0 & 88.0 & -- & \textbf{98.0} \\
PlateToBowl                & 57.0 & 52.0 & 50.0 & 54.0 & 48.0 & -- & \textbf{68.0} \\
TrayToPlate                & 71.0 & 64.0 & 64.0 & 68.0 & 88.0 & -- & \textbf{90.0} \\
\multicolumn{8}{c}{\ldots\,(24 tasks total; see Appendix for full results)} \\
\hline
\textbf{Average (24 tasks)} & 47.6 & 43.9 & 55.0 & 58.3 & 63.2 & 70.2 & \textbf{72.8} \\
\hline
\end{tabular*}
\end{table*}

As shown in Table~\ref{tab:robocasa_gr1_results}, \Ours achieves \textbf{72.8\%} average success rate, surpassing all baselines including DIAL~\cite{chen2026dial} (70.2\%), JoyAI-RA~\cite{zhang2026joyaira} (63.2\%), ABot-M0~\cite{yang2026abot} (58.3\%), and FLARE~\cite{pmlr-v305-zheng25a} (55.0\%). The gains are consistent across both articulated-object interaction and pick-and-place rearrangement task categories, suggesting that the camera-space action interface and reliability-aware training generalize broadly rather than benefiting a narrow subset of tasks.

\subsubsection{RoboTwin~2.0}
\label{sec:robotwin}

RoboTwin~2.0 is a bimanual tabletop manipulation benchmark covering 50 tasks with strong domain randomization. 
We train on 2,500 clean demonstrations (50 per task) plus 25,000 randomized demonstrations (500 per task), and evaluate under both Easy/Clean and Hard/Randomized settings. Overall results are shown in Table~\ref{tab:robotwin2_results}, with full per-task results in Appendix~\ref{app:full-robotwin-results}, Table~\ref{tab:robotwin2_full_task_results}.

\begin{table*}[t]
\centering
\caption{Overall evaluation results on the RoboTwin~2.0 benchmark. Success rates (\%) averaged over 50 tasks, 100 trials per task. Easy denotes the clean setting and Hard denotes the randomized setting. Full per-task results are in Appendix~\ref{app:full-robotwin-results}.}
\label{tab:robotwin2_results}
\vspace{0.3em}
\small
\renewcommand{\arraystretch}{1.1}
\setlength{\tabcolsep}{3pt}
\begin{tabular*}{\textwidth}{@{\extracolsep{\fill}}lcccccccccccccc}
\hline
\textbf{Method} & \multicolumn{2}{c}{$\pi_{0.5}$} & \multicolumn{2}{c}{Motus} & \multicolumn{2}{c}{LingBot-VLA} & \multicolumn{2}{c}{ABot-M0} & \multicolumn{2}{c}{JoyAI-RA} & \multicolumn{2}{c}{Hy-VLA} &\multicolumn{2}{c}{\Ours} \\
 & Easy & Hard & Easy & Hard & Easy & Hard & Easy & Hard & Easy & Hard & Easy & Hard & Easy & Hard \\
\hline
Success Rate & 82.74 & 76.76 & 88.66 & 87.02 & 88.56 & 86.68 & 86.06 & 85.08 & 90.48 & 89.28 & 90.9 & 90.1 & \textbf{91.12} & \textbf{90.62} \\
\hline
\end{tabular*}
\end{table*}

\Ours achieves \textbf{91.12\%} average success rate on the Easy/Clean setting and \textbf{90.62\%} on the Hard/Randomized setting, surpassing JoyAI-RA by 0.64\% and 1.34\%, respectively.
The improvement is distributed across diverse manipulation primitives---grasping, placement, tool use, and bimanual coordination---indicating that the unified pretraining recipe transfers effectively to multi-task bimanual control under strong domain randomization.

\subsection{Real-Robot Evaluation}
\label{sec:real_robot}

\begin{figure}[t]
  \centering
  \begin{subfigure}[t]{0.48\linewidth}
    \centering
    \includegraphics[width=\linewidth]{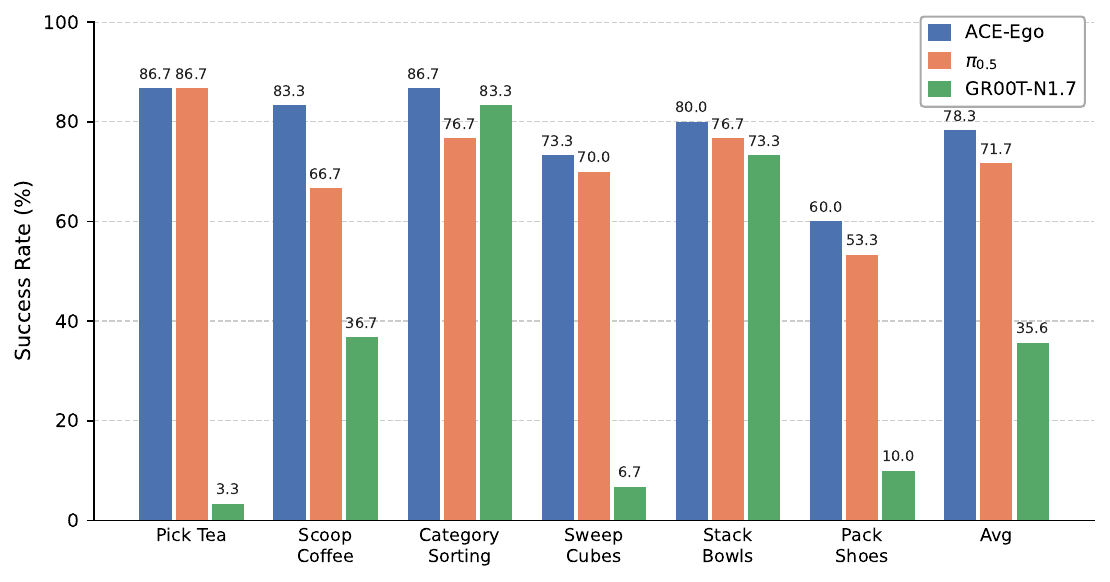}
    \caption{Real-robot results on the ARX bimanual platform vs.\ $\pi_{0.5}$. Trials: 30 per task.}
    \label{fig:real_robot}
  \end{subfigure}
  \hfill
  \begin{subfigure}[t]{0.48\linewidth}
    \centering
    \includegraphics[width=\linewidth]{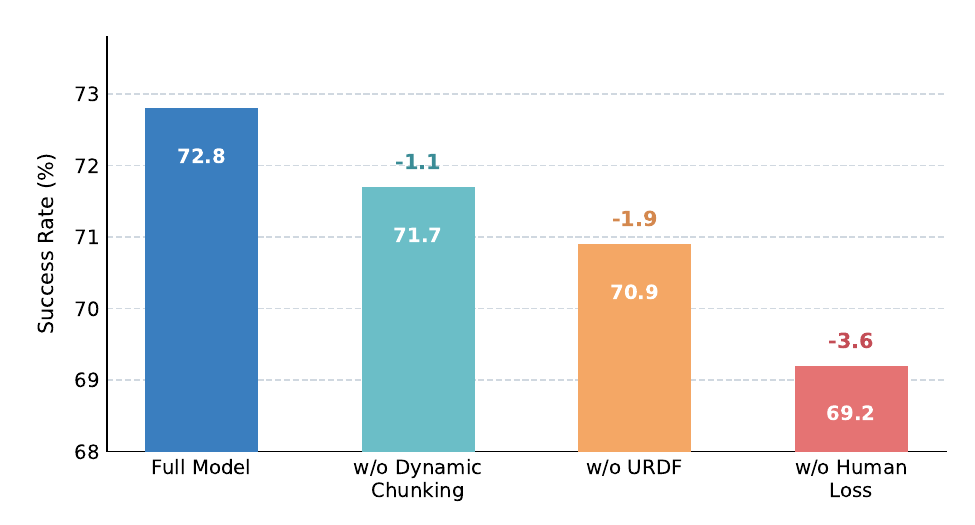}
    \caption{Component ablation on RoboCasa GR1 TableTop. Each bar shows the effect of removing one component from the full model.}
    \label{fig:ablation}
  \end{subfigure}
  \caption{Real-robot evaluation (a) and ablation study (b) for \Ours.}
  \label{fig:real_robot_ablation}
\end{figure}

We evaluate \OURS on an ARX bimanual platform equipped with a head-mounted RGB-D camera, controlled via camera-space delta end-effector commands. 
The policy outputs actions directly in the head-camera coordinate frame and is deployed by simply applying a single camera extrinsic at inference time.

We evaluate on six manipulation tasks of increasing complexity—spanning single-arm pick-and-place, long-horizon multi-step manipulation, contact-rich bimanual coordination, and language-grounded semantic reasoning (see Appendix~\ref{app:real-robot-task-descriptions} for full descriptions). 
We compare against two strong baseline methods: $\pi_{0.5}$ fine-tuned on the same downstream task data, and GR00T-N1.7.
A trial is considered successful only if the robot completes the entire task sequence without human intervention; per-task success criteria are detailed in Appendix~\ref{app:real-robot-success-criteria}, and the quantitative results are summarized in Figure~\ref{fig:real_robot_ablation}(a).

\Ours achieves a \textbf{78.3\%} average success rate across the six tasks, outperforming $\pi_{0.5}$ (71.7\%) by 6.6\%, and demonstrating a decisive margin over GR00T-N1.7, which struggles on several long-horizon sequences and obtains a 35.6\% average success rate.
Specifically, \Ours leads across five out of the six tasks. On Scoop Coffee, a contact-rich bimanual task requiring tight spatiotemporal coordination between both arms, \Ours achieves 86.7\%, outperforming $\pi_{0.5}$ (70.0\%) by 16.7\% and GR00T-N1.7 (36.7\%) by 50.0\%.

In the multi-class object placement task, Category Sorting, \Ours maintains a steady performance of 90.0\%, compared to 80.0\% for $\pi_{0.5}$ and 83.3\% for GR00T-N1.7. While GR00T-N1.7 exhibits reasonable capability on relatively structured setups such as Stack Bowls (73.3\%), its execution consistency drops sharply on tasks that require extended horizontal trajectories or explicit bimanual coordination, such as Sweep Cubes (6.7\%).

In sharp contrast, \Ours demonstrates its clear advantage in spatiotemporal alignment on Scoop Coffee, a contact-rich bimanual task requiring tight synchronization between both arms, sustaining an 86.7\% success rate while GR00T-N1.7 falls to 36.7\%. On Pack Shoes, which features the longest operational sequence including a delicate lid-closing phase, all evaluated models experience a visible degradation in performance. This joint performance drop suggests that managing compounding trajectory drift over long-horizon manipulation chains remains a common shared challenge for existing pretrained VLA architectures.

\subsection{Ablation Studies}
\label{sec:ablation_studies}

\paragraph{Component ablation.}
We ablate three components of \Ours on RoboCasa GR1 TableTop, removing one at a time from the full model and reporting the average success rates of the checkpoints trained for 190K steps (Figure~\ref{fig:real_robot_ablation}(b)).
Removing either one of the three components reduces performance.
Removing morphology tokens makes the success rate drop from 72.8\% to 70.9\% ($-1.9\%$): even though all sources share the same camera-space action format, different robot platforms have different kinematic structures, and the morphology tokens provide the action expert kinematics-related information.
Removing time-aligned action chunking drops the success rate to 71.7\% ($-1.1\%$): a fixed number of actions now covers different physical durations across datasets collected at different frame rates and introduces temporal inconsistency between the mixed-source data.
Removing the reliability-aware human auxiliary loss leads to the largest success rate drop to 69.2\% ($-3.6\%$): without label-quality weighting, noisy pseudo-actions from human videos receive equal supervision weight as sensor-logged robot actions, which confuses the action expert training.

\paragraph{Data source ablation.}
We also evaluate three pretraining configurations on RoboCasa GR1 TableTop to assess the contribution of each data source (see Table~\ref{tab:human_scaling}).

\begin{table}[h]
  \centering
  \small
  \caption{Pretraining data ablation on RoboCasa GR1 TableTop (success rate).}
  \label{tab:human_scaling}
  \begin{tabular}{p{0.45\textwidth}r}
    \toprule
    Pretraining Configuration & Success Rate (\%) \\
    \midrule
    Robot + Human (full \Ours)          & \textbf{72.8} \\
    Robot Only (no human video)         & 68.3 \\
    From Qwen (no embodied pretrain)    & 65.4 \\
    \bottomrule
  \end{tabular}
\end{table}

The success rate increases with each additional data source.
The Qwen-initialized model without embodied pretraining reaches 65.4\%.
Adding robot data raises the success rate to 68.3\% ($+2.9\%$), showing that embodied pretraining provides action-level knowledge that pure language-vision pretraining does not.
Adding human videos further raises the rate to 72.8\% ($+4.5\%$, the largest single gain), showing that human videos contribute diverse behavioral coverage beyond the robot demonstrations alone.


\subsection{Human Data for Augmented Fine-Tuning}
\label{sec:human-mid-training}

We investigate how human egocentric videos improve task-specific adaptation when robot demonstration data alone are insufficient.
Starting from the pretrained \Ours checkpoint, we fine-tune on the Sweep Cubes task using only 34 robot demonstrations (2 sessions, $\sim$45.8K frames). With robot data alone, the policy achieves only a 10\% success rate (1/10 trials).

\begin{figure}[t]
  \centering
  \includegraphics[width=\linewidth]{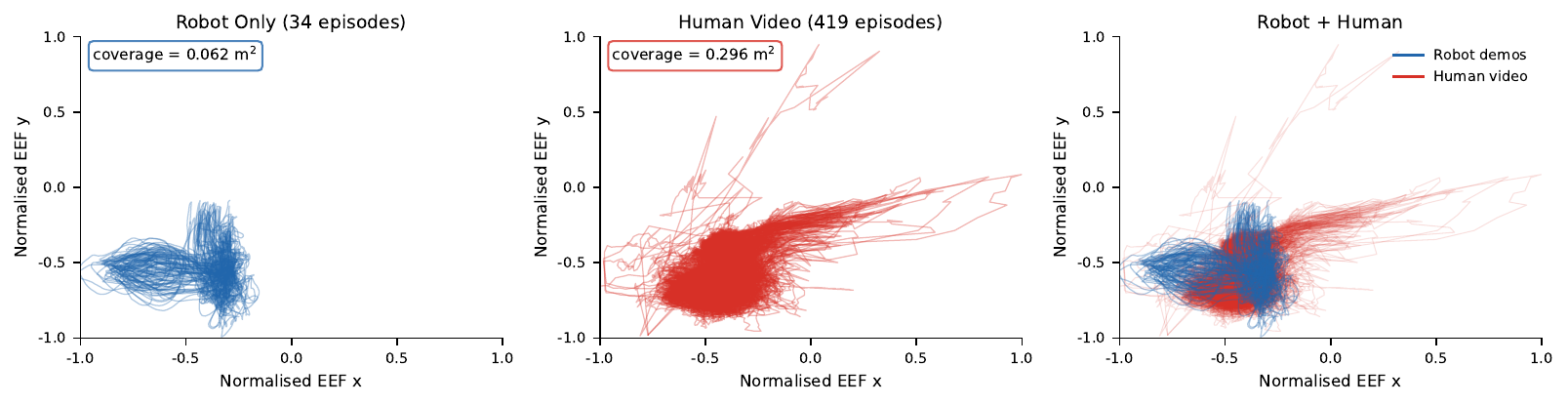}
  \caption{Right end-effector trajectories for the Sweep Cubes fine-tuning data, projected onto the horizontal plane. Axes are remapped to $[-1, 1]$ for readability; each panel uses the same scale.
  \textbf{Left:} 34 robot demonstrations are concentrated in a small region (0.062 m$^2$ convex-hull area).
  \textbf{Middle:} 419 human video episodes cover a substantially broader area (0.296 m$^2$, 4.8$\times$ larger).
  \textbf{Right:} both sources overlaid, showing the robot cluster embedded within the wider human distribution.}
  \label{fig:action_distribution}
\end{figure}

Figure~\ref{fig:action_distribution} shows the reason: The 34 robot demonstrations occupy only a narrow region of the action space, covering only 0.062 m$^2$ of the end-effector workspace. The 419 episodes of task-matched human video spread across 0.296 m$^2$, 4.8$\times$ broader coverage, filling in motion patterns that the limited robot data does not include.
Augmenting the fine-tuning mixture with this human video ($\sim$117.5K frames) increases the success rate to 40\% (4/10 trials), a 4$\times$ improvement, confirming that human video provides complementary action coverage and substantially recovers performance in data-scarce fine-tuning regimes.

\section{Conclusion}
\label{sec:conclusion}
We presented \Ours, a VLA pretraining framework that jointly resolves representational and label-quality heterogeneity when learning from large-scale human and multi-embodiment robot data.
\Ours resolves representation heterogeneity through a unified action representation, aligning heterogeneous sources along spatial, structural, and temporal spaces via camera-space actions, cross-embodiment morphology tokens, and time-aligned action chunking. It further addresses the supervision-quality mismatch through a reliability-aware training objective that provides noise-resilient supervision for large-scale pseudo-action labels.
Instantiated on a 6.0K+ hour pool spanning multiple robot platforms, simulation environments, and 1.48K hours of large-scale egocentric human video, \Ours achieves 72.8\% on RoboCasa GR1 TableTop and 91.12\%/90.62\% on RoboTwin~2.0 Easy/Hard splits, outperforming all compared methods; human-augmented fine-tuning further demonstrates a 4$\times$ improvement in data-scarce regimes.
On a real bimanual ARX platform, \Ours reaches a 78.3\% average success rate across six physical manipulation tasks, consistently outperforming fine-tuned $\pi_{0.5}$ (71.7\%) and demonstrating a decisive margin over GR00T-N1.7 (35.6\%), with prominent capabilities in multi-step sequential execution and coordinated bimanual control.

\section{Limitations} 
While \Ours demonstrates strong performance across simulation and real-world benchmarks, several directions remain open. Our current evaluation focuses on tabletop manipulation; extending to mobile manipulation, whole-body humanoid control, or deformable-object tasks would test the generality of the camera-space action interface under more diverse spatial conventions and longer task horizons. The pretraining pool, though large, does not yet include dexterous hand data or force/torque sensing; incorporating richer modalities could further improve contact-rich manipulation. Finally, scaling the human-video portion and improving the fidelity of pseudo-action pipelines---particularly for rotation and fine-grained finger motion---would allow the reliability-aware objective to supervise additional action dimensions beyond position, potentially unlocking stronger transfer from human demonstrations to robot control.


\clearpage
\acknowledgments{If a paper is accepted, the final camera-ready version will (and probably should) include acknowledgments. All acknowledgments go at the end of the paper, including thanks to reviewers who gave useful comments, to colleagues who contributed to the ideas, and to funding agencies and corporate sponsors that provided financial support.}


\bibliography{main}  
\clearpage
\appendix

\section{Additional Method Details}
\label{app:method-details}
\begin{figure}[htbp]
  \centering
  \includegraphics[width=0.95\linewidth]{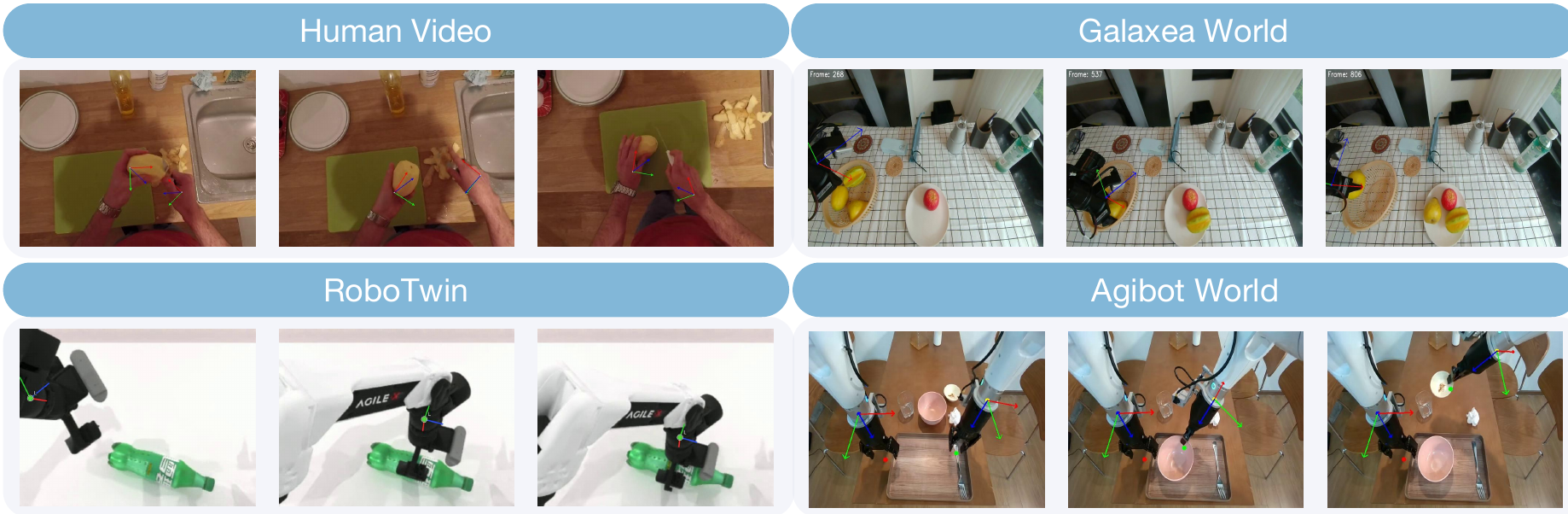}
  \caption{Camera-space action visualization across real robot demonstrations, simulation rollouts, and human egocentric video. All sources express end-effector or hand motion relative to the head-camera frame, making heterogeneous action labels comparable under the same observation-aligned coordinate convention.}
  \label{fig:camera-space-action-vis}
\end{figure}

\subsection{Camera-Space Action Standardization and Layout}
\label{app:robot-action-standardization}

This subsection details the spatial alignment pipeline for both robot and human sources introduced in Sec.~\ref{sec:camera-space-action}.

\paragraph{Coordinate Transformation.}
For robot sources, action labels are sensor-grounded end-effector poses. If an end-effector pose is reported in a source frame $s$ (e.g., robot base or world frame), we convert it to the head-camera frame using the calibrated camera extrinsic as formulated in Eq.~\ref{eq:cam-transform}. For bimanual platforms, this transformation is applied independently to the left and right end-effectors. If a dataset already stores aligned camera-frame actions, the conversion is the identity.

\paragraph{Continuous 6D Orientation.}
To avoid the discontinuities of quaternions or Euler angles during training, orientations are normalized to a continuous 6D representation~\cite{zhou2019continuity}. Quaternions are first converted to rotation matrices $R_{\mathrm{cam},ee} \in \mathrm{SO}(3)$. We then extract and concatenate the first two columns of the rotation matrix:
\begin{equation}
    \mathrm{rot6d}(R_{\mathrm{cam},ee})
    =
    \left[
    R_{\mathrm{cam},ee}^{(:,1)};\,
    R_{\mathrm{cam},ee}^{(:,2)}
    \right] \in \mathbb{R}^6.
\end{equation}

\paragraph{Human Hand-Centric Frame Derivation.}
To parameterize human trajectories into the identical coordinate space, we construct a stable hand-centric coordinate frame $R_{\mathrm{cam}, hand} = [\mathbf{x}, \mathbf{y}, \mathbf{z}] \in \mathrm{SO}(3)$ from the reconstructed hand mesh. We designate the wrist joint $\mathbf{p}_{\mathrm{wrist}}$ as the origin. Let $\mathbf{p}_{\mathrm{palm}}$ denote the palm centroid, computed as the mean of the index, middle, and ring fingertip positions. The orthogonal axes of the hand frame are constructed as:
\begin{equation}
  \mathbf{x} = \frac{\mathbf{p}_{\mathrm{palm}} - \mathbf{p}_{\mathrm{wrist}}}{\|\mathbf{p}_{\mathrm{palm}} - \mathbf{p}_{\mathrm{wrist}}\|_2},
  \quad
  \mathbf{z} = \hat{n}(\mathbf{p}_{\mathrm{wrist}}, \mathbf{p}_{\mathrm{thumb}}, \mathbf{p}_{\mathrm{middle}}),
  \quad
  \mathbf{y} = \mathbf{z} \times \mathbf{x},
\end{equation}
where $\hat{n}(\mathbf{a},\mathbf{b},\mathbf{c})$ denotes the unit normal of the plane defined by points $\mathbf{a},\mathbf{b},\mathbf{c}$, with the sign chosen to point away from the palm. The resulting rotation matrix $R_{\mathrm{cam}, hand}$ is then converted to the continuous 6D representation. For gripper openness, the thumb-to-palm distance $d_t = \|\mathbf{p}_{\mathrm{thumb},t} - \mathbf{p}_{\mathrm{palm},t}\|_2$ is linearly normalized to the gripper stroke range of our robot platforms.

\paragraph{Unified 22-Dimensional Action Layout.}
After standardization, both robot and human trajectories are mapped into a unified 22-dimensional bimanual action vector $\mathbf{a} \in \mathbb{R}^{22}$. The vector is structured as a concatenation of symmetric 11-dimensional single-arm action blocks:
\begin{equation}
\mathbf{a} = \left[ \mathbf{a}_{\mathrm{left}};\, \mathbf{a}_{\mathrm{right}} \right] \in \mathbb{R}^{22},
\end{equation}
where each arm's action block $\mathbf{a}_{\mathrm{arm}} \in \mathbb{R}^{11}$ is defined as:
\begin{equation}
\mathbf{a}_{\mathrm{arm}} = \left[ \underbrace{p_x, p_y, p_z}_{\text{Position (3D)}}, \, \underbrace{r_1, \dots, r_6}_{\text{Continuous Orientation (6D)}}, \, \underbrace{g}_{\text{Gripper (1D)}}, \, \underbrace{\alpha}_{\text{Activity Flag (1D)}} \right].
\end{equation}
The binary activity flag $\alpha \in \{0, 1\}$ indicates whether the corresponding arm is active in the dataset, allowing the policy to seamlessly handle both single-arm and bimanual embodiments.

\paragraph{Projection-Based Data Validation.}
We utilize camera projection as a validation signal to filter out tracking failures. Given a camera-frame end-effector position $(X,Y,Z)$ and camera intrinsics $(f_x,f_y,c_x,c_y)$, the projection onto the image plane is:
\begin{equation}
    u = f_x \frac{X}{Z} + c_x,
    \qquad
    v = f_y \frac{Y}{Z} + c_y.
\end{equation}
Frames with non-positive depth ($Z \le 0$) or projections falling outside the image boundaries are flagged and masked out using the action validity mask $M$.

\subsection{Robot Kinematic Graph Construction}
\label{app:urdf-graph-construction}

For each robot embodiment with a URDF, we build the compact kinematic graph $\mathcal{G}_{r}$ used by the morphology encoder. This graph feeds only the morphology conditioning and never enters the shared vision-language trunk. Training samples carry a robot identifier rather than a URDF path. A registry maps each identifier to a canonical robot name, a URDF file, a base link, and left/right end-effector links. We cache the graph tensors by canonical name, so URDF parsing and graph construction happen once and are reused across training.

We use a joint-centric graph: each node is a URDF joint, and edges follow the parent--child relations of the kinematic tree. This puts the quantities most relevant to control---joint type, motion axis, origin transform, limits, and actuation state---directly on the nodes. The registered end-effector links define the left and right manipulation chains $\mathcal{C}^{r}_{L}$ and $\mathcal{C}^{r}_{R}$, which we use to mark action-relevant joints and to measure each joint's distance to the terminal joints of the two chains.

Each joint carries a fixed-dimensional descriptor with four groups of information: local kinematic attributes, range and actuation properties, graph-topological position, and relation to the left/right end-effector chains. In our implementation this is a $29$-dimensional node feature vector. Stacking the descriptors for all joints gives a node feature matrix $X_r \in \mathbb{R}^{N_r \times 29}$ for robot $r$, where $N_r$ is the number of URDF joints. The cached payload holds $X_r$, the normalized adjacency matrix, the left/right chain masks $\mathcal{C}^{r}_{L}, \mathcal{C}^{r}_{R}$, and joint metadata.

\subsection{Morphology Encoder}
\label{app:morphology-encoder-details}

This subsection details the URDF encoder $E_{\mathrm{urdf}}$ introduced in Sec.~\ref{sec:urdf-conditioning}. It maps the cached graph $\mathcal{G}_r$ (Appendix~\ref{app:urdf-graph-construction}) to the body summary $z_{\mathrm{body}}^{r}$ and manipulation-chain summary $z_{\mathrm{chain}}^{r}$ that make up $E_{\mathrm{urdf}}(\mathcal{G}_r)$ in Eq.~\ref{eq:morph-token}.

The encoder has two stages: it first contextualizes each joint by message passing over the kinematic tree, then pools the joint states into the two summaries.

\paragraph{Message Passing.}
Starting from the joint descriptors $X_r$, the encoder runs $L$ residual layers:
\begin{equation}
    H^{(0)} = \phi_{\mathrm{in}}(X_r),
    \qquad
    H^{(\ell+1)}
    =
    H^{(\ell)}
    +
    \phi_{\ell}
    \!\left(
    \left[
    H^{(\ell)};\,
    \bar{A}_r H^{(\ell)}
    \right]
    \right),
    \quad \ell = 0,\dots,L-1,
\end{equation}
where $\bar{A}_r = D_r^{-1}(A_r + I)$ is the adjacency matrix with self-loops, row-normalized by its degree matrix $D_r$; $\phi_{\mathrm{in}}$ and $\phi_{\ell}$ are MLPs. At each layer, $\bar{A}_r H^{(\ell)}$ averages every joint with its neighbors, while the residual path preserves the joint's own state.

\paragraph{Pooling and Concatenation.}
Writing $\operatorname{mp}(\mathcal{S}) = \frac{1}{|\mathcal{S}|}\sum_{j \in \mathcal{S}} H^{(L)}_j$ for the mean of the final states over a joint set $\mathcal{S}$, the two summaries are:
\begin{equation}
    z_{\mathrm{body}}^{r} = \rho_{\mathrm{body}}\!\left(\operatorname{mp}(\mathcal{J}_r)\right),
    \qquad
    z_{\mathrm{chain}}^{r} = \rho_{\mathrm{chain}}\!\left(\left[\operatorname{mp}(\mathcal{C}^{r}_{L});\,\operatorname{mp}(\mathcal{C}^{r}_{R})\right]\right),
\end{equation}
where $\mathcal{J}_r$ is the full joint set and $\mathcal{C}^{r}_{L}, \mathcal{C}^{r}_{R}$ are the left and right end-effector chains from Appendix~\ref{app:urdf-graph-construction}, and $\rho_{\mathrm{body}}, \rho_{\mathrm{chain}}$ are MLPs. The body summary captures the global embodiment, while the chain summary focuses on the kinematic paths most involved in manipulation. The final URDF representation is the concatenation of these two summaries:
\begin{equation}
E_{\mathrm{urdf}}(\mathcal{G}_r) = \left[ z_{\mathrm{body}}^{r};\, z_{\mathrm{chain}}^{r} \right],
\end{equation}
which is then projected by $P_{\mathrm{morph}}$ into the shared morphology token space as shown in Eq.~\ref{eq:morph-token}.

\subsection{Human Surrogate Morphology Embeddings}
\label{app:human-surrogate-details}

Human egocentric video has no robot URDF, so the encoder $E_{\mathrm{urdf}}$ of Appendix~\ref{app:morphology-encoder-details} does not apply. Human sources still differ from one another in embodiment and capture conditions, and the action expert should be conditioned on these differences just as it is for robots. We therefore represent each human-video source by a learned surrogate embedding $e_{d} \in \mathbb{R}^{D}$ and project it with $P_{\mathrm{surr}}$ into the same morphology token as the URDF-conditioned robots (Eq.~\ref{eq:morph-token}).

The surrogate absorbs stable source-level factors that the shared camera-space action representation does not explain: camera placement and field of view, the visual domain of each corpus, annotation quality, and source-specific action statistics. These factors stay roughly constant within a source but differ across sources, so a per-source embedding fits them better than a per-sample input. We can allocate the embedding per dataset (one $e_{d}$ per human-video source) or share it across all human-video data, and we use the per-dataset variant by default. After the morphology-token interface, the action expert treats robot and human-video samples the same way: robots get the condition from a structured URDF graph, and human-video sources get it from learned surrogate embeddings. The surrogate embeddings $e_d$ are randomly initialized and optimized end-to-end during pretraining alongside all other model parameters.





\subsection{Reliability-Aware Human Auxiliary Loss Details}
\label{app:reliability-aware-human-details}

This subsection expands the mathematical formulation of the spatiotemporal reliability weight $W_{t,j}$ and the temporal smoothing summarized in Sec.~\ref{sec:reliability-aware-objective}. All quantitative thresholds are collected in Table~\ref{tab:reliability-hparams}.

\paragraph{Hierarchical Reliability Decomposition.}
As introduced in Eq.~\ref{eq:reliability}, the spatiotemporal reliability $W_{t,j}$ of channel $j$ at step $t$ is decomposed into a static channel-level prior $\rho_j$ and a dynamic step-level weight $w_{t,j}$. We further decompose the step-level weight $w_{t,j}$ into a dataset-level prior $w_{\mathrm{data}}$ and a local step-level smoothness factor $w_{\mathrm{step}}$, yielding the final hierarchical formulation:
\begin{equation}
W_{t,j} = \rho_j \cdot w_{\mathrm{data}}(d, h(j)) \cdot w_{\mathrm{step}}(t, h(j)),
\end{equation}
where $h(j)$ maps action channel $j$ to its corresponding hand (left or right). The dataset prior $w_{\mathrm{data}}$ sets a global quality ceiling for each source, while the step weight $w_{\mathrm{step}}$ modulates it locally in response to tracking anomalies.

\paragraph{Normalization.}
The human auxiliary loss $\mathcal{L}_{\mathrm{haux}}$ is normalized per sample by the total effective supervision weight:
\begin{equation}
Z = \sum_{t,j} M_{t,j}\,W_{t,j}.
\end{equation}
This formulation ensures that the auxiliary loss is scale-invariant to the number of valid entries and concentrates supervision on highly reliable channels. When a minibatch contains no human sample, $\mathcal{L}_{\mathrm{haux}}$ is set to zero.


\paragraph{Step-Level Smoothness Weight.}
The time-step factor $w_{\mathrm{step}}$ down-weights segments whose motion is locally implausible, which typically indicates reconstruction error rather than genuine fast motion. For hand $h$ we compute, from the clean position chunk, the first- and second-order differences:
\begin{equation}
\Delta p_t^{h} = \left\|p_t^{h} - p_{t-1}^{h}\right\|_2,
\qquad
\Delta^2 p_t^{h} =
\left\|p_{t+1}^{h} - 2p_t^{h} + p_{t-1}^{h}\right\|_2,
\end{equation}
which measure inter-frame speed and jerk, respectively. For each human-video dataset $d$ and hand $h$, we precompute robust thresholds $\tau_{\mathrm{jump}}(d,h)$ and $\tau_{\mathrm{jerk}}(d,h)$ as the 95th percentiles of $\Delta p_t^h$ and $\Delta^2 p_t^h$ over clean position chunks from that dataset. At training time, we compute:
\begin{equation}
q_{t,h}
=
\max\!\left(
\frac{\Delta p_t^h}{\tau_{\mathrm{jump}}(d,h)},
\frac{\Delta^2 p_t^h}{\tau_{\mathrm{jerk}}(d,h)}
\right).
\end{equation}
The step weight is then formulated as:
\begin{equation}
w_{\mathrm{step}}(t,h)
=
\begin{cases}
1, & q_{t,h} \le 1,\\
\max\!\left\{w_{\min}, \exp[-\alpha(q_{t,h}-1)]\right\}, & q_{t,h} > 1.
\end{cases}
\end{equation}
Thus, nominally smooth steps retain full weight, while unusually large jumps or jerks relative to the dataset-hand statistics are softly attenuated.

\paragraph{Dataset-Level Prior.}
Each human-video source carries a different reconstruction quality, so we attach a per-source, per-hand prior $w_{\mathrm{data}}(d,h) \in (0,1]$. For dataset $d$ and hand $h$ this prior is estimated from the clean position trajectories of that source: we aggregate the fraction of frames surviving the sanity filters together with the median normalized jerk of the retained trajectories, and map sources with higher survival and lower jerk to priors closer to $1$. The prior is computed once per source and held fixed during training.

\paragraph{Temporal Smoothing.}
Before constructing the auxiliary target velocity $(\tilde{\mathbf{a}} - \boldsymbol{\epsilon})$, we apply a temporal smoothing filter of window $W_{\mathrm{smooth}}$ to the clean human action targets. This suppresses high-frequency pose jitter introduced by per-frame hand mesh regression without altering the supervised dimensions or the $W_{t,j}$ weights, which are computed from the pre-smoothing chunk.

\begin{table}[htbp]
\centering
\small
\caption{Reliability-aware human supervision hyperparameters used in the
human auxiliary loss (Section~\ref{sec:human-auxiliary-loss}). Values are
shared across the six human-video sources unless noted otherwise.}
\label{tab:reliability-hparams}
\renewcommand{\arraystretch}{1.1}
\begin{tabular}{lll}
\toprule
\textbf{Component} & \textbf{Hyperparameter} & \textbf{Value} \\
\midrule
\multirow{2}{*}{Auxiliary loss}
  & Loss weight $\lambda_{\mathrm{haux}}$           & 0.1 \\
  & Huber transition $\beta$                        & 1.0 \\

\midrule
\multirow{3}{*}{Human channel prior $\rho_j$}
  & Position channels $\mathcal{P}$ ($\rho_j$)        & 1.0 \\
  & Rotation / gripper channels ($\rho_{\mathrm{low}}$) & 0.001 \\  
  & Position channel set $\mathcal{P}$                  & wrist xyz, both hands (6 dims) \\
\midrule
\multirow{4}{*}{Step weight}
  & Jump threshold $\tau_{\mathrm{jump}}(d,h)$      & per-dataset/hand 95th percentile \\
  & Jerk threshold $\tau_{\mathrm{jerk}}(d,h)$      & per-dataset/hand 95th percentile \\
  & Attenuation sharpness $\alpha$                  & 1.5 \\
  & Minimum step weight $w_{\min}$                  & 0.2 \\
\midrule
\multirow{2}{*}{Dataset prior}
  & Prior range $w_{\mathrm{data}}$                 & [0.25, 1.0] \\
  & Estimation                                      & q95 jump/jerk ratio to robot reference \\
\midrule
Smoothing
  & Smoothing window $W_{\mathrm{smooth}}$          & 3 frames \\
\bottomrule
\end{tabular}
\end{table}

\section{Training Details}
\label{app:training-details}

\subsection{Architecture, Training, and Evaluation Protocol}
\label{app:training-details-protocol}

\paragraph{Model architecture.}
\Ours uses Qwen3-VL-4B-Instruct as the vision-language backbone and a
flow-matching Diffusion Transformer ($\sim$600M parameters) as the action
expert. Images from head and wrist cameras are processed at $256{\times}256$
resolution, and actions are decoded in 4 flow-matching steps at inference.
Full layer and dimension configurations are listed in
Table~\ref{tab:architecture}.

\paragraph{Training protocol.}
Pretraining runs on 128$\times$A800 (80GB) GPUs with AdamW and a
cosine schedule; task-specific fine-tuning uses 16$\times$A800 GPUs with the
same optimizer settings. Full optimizer hyperparameters, learning rates, and
schedule are listed in Table~\ref{tab:hyperparams}.

\paragraph{Evaluation protocol.}
RoboCasa evaluates 50 rollouts per task across 24 tasks. RoboTwin 2.0
evaluates 100 trials per task across 50 tasks under both Easy and Hard
settings. Real-robot experiments use 30 trials per task.
A trial is considered successful only if the robot completes the entire task
sequence without human intervention; per-task real-robot success criteria are
detailed in Appendix~\ref{app:real-robot-success-criteria}.

\begin{table}[htbp]
\centering
\small
\caption{Model architecture configuration for \Ours.}
\label{tab:architecture}
\renewcommand{\arraystretch}{1.1}
\begin{tabular}{ll}
\toprule
\textbf{Component} & \textbf{Configuration} \\
\midrule
VLM backbone              & Qwen3-VL-4B-Instruct ($\sim$4B params) \\
\quad Vision encoder      & 24 layers, patch size $16{\times}16$ \\
\quad Language model      & 36 layers, hidden size 2560 \\
Input resolution          & $256{\times}256$ (head + wrist) \\
\midrule
Action expert             & Flow-matching DiT \\
\quad Layers / hidden     & 36 / 1024 \\
\quad Attention heads     & 16 (head dim 64) \\
\quad Parameters          & $\sim$600M \\
Inference decoding steps  & 4 \\
\bottomrule
\end{tabular}
\end{table}

\subsection{Hyperparameters}
\label{app:hyperparameters}

Table~\ref{tab:hyperparams} summarizes the key training hyperparameters for \Ours pretraining and fine-tuning.

\begin{table}[htbp]
\centering
\small
\caption{Training hyperparameters for \Ours pretraining and fine-tuning.}
\label{tab:hyperparams}
\begin{tabular}{ll}
\toprule
\textbf{Hyperparameter} & \textbf{Value} \\
\midrule
VLM backbone & Qwen3-VL-4B-Instruct \\
Action expert & Flow-matching DiT (36 layers, 1024 hidden, 16 heads) \\
Action expert parameters & $\sim$600M \\
Image resolution & 256$\times$256 \\
Action horizon & dataset-specific \(H_d = \mathrm{round}(f_d T^\star)\), with \(T^\star = 2\,\mathrm{s}\) \\ & 40 steps for 20 Hz RoboCasa SFT sources \\
Flow matching inference steps & 4 \\
Optimizer & AdamW ($\beta_1$=0.9, $\beta_2$=0.95, $\epsilon$=1e-8) \\
VLM learning rate & 2e-5 \\
Action expert learning rate & 1e-4 \\
LR schedule & Cosine with min LR 5e-7 \\
Warmup steps & 5000 \\
Weight decay & 1e-8 \\
Gradient clipping & 1.0 \\
Batch size (per device) & 8 \\
Pretraining GPUs & 128$\times$A800 (80GB) \\
Pretraining steps & 200K \\
Fine-tuning GPUs & 16$\times$A800 (80GB) \\
$\lambda_{\mathrm{haux}}$ & 0.1 \\
Human quality mode & dataset + step \\
Repeated diffusion steps & 4 \\
\bottomrule
\end{tabular}
\end{table}

\subsection{Dataset Mixtures and Sampling}
\label{app:dataset-mixtures}

Full dataset statistics are reported in Table~\ref{tab:pretraining-data-main} (main text). The pool is assembled from named dataset groups rather than from one monolithic corpus, which lets us
control the sampling weight and preprocessing path of each source independently. The Ego4D entry combines cooking and non-cooking splits; hours are computed from LeRobot metadata as
frames/(fps$\times$3600).

Sampling is performed at the dataset-group level. Each mixture entry has a sampling weight and a source type. Human-video sources are marked with the \texttt{human\_video} source type and are routed
through the camera-space pseudo-action path and reliability-aware human loss. Robot sources use their corresponding robot type and are supervised with the main robot action objective. This separation lets
large but noisy human-video corpora contribute broad visual and behavioral coverage without overwhelming higher-fidelity robot demonstrations.

\section{Additional Experiments}
\label{app:additional-experiments}



\subsection{Real-Robot Task Descriptions}
\label{app:real-robot-task-descriptions}

The six real-robot tasks, ordered by increasing complexity, are:

\begin{itemize}
    \item \textbf{Pick Tea}: grasp a shopping basket and place it at the workspace center, then pick up a tea box and drop it into the basket.
    \item \textbf{Scoop Coffee}: the right arm grasps a coffee scoop while the left arm holds a coffee canister; the right arm scoops coffee from the canister and pours it into a designated cup.
    \item \textbf{Category Sorting}: multiple objects (toiletries and beverages) are scattered on the workspace; the robot sorts each object into the corresponding bin based on semantic category.
    \item \textbf{Sweep Cubes}: the left arm holds a dustpan in a fixed pose while the right arm uses a broom to sweep cubes on the workspace into the dustpan.
    \item \textbf{Stack Bowls}: sequentially pick up three bowls from the workspace and stack them vertically.
    \item \textbf{Pack Shoes}: move a shoe box to the workspace center, sequentially place two shoes inside, and close the lid.
\end{itemize}

\subsection{Real-Robot Success Criteria}
\label{app:real-robot-success-criteria}

Each real-robot trial is evaluated by a human judge. A trial is marked successful only if the robot completes the full task sequence without human intervention. The per-task success definitions are:

\begin{itemize}
    \item \textbf{Pick Tea}: The shopping basket is placed at the workspace center and the tea box is dropped inside the basket.
    \item \textbf{Scoop Coffee}: Coffee is scooped from the canister and a visible amount is deposited into the designated cup.
    \item \textbf{Category Sorting}: All scattered objects are placed into their correct category bins (toiletries vs.\ beverages).
    \item \textbf{Sweep Cubes}: All cubes on the workspace are swept into the dustpan held by the left arm.
    \item \textbf{Stack Bowls}: All three bowls are picked up and stacked vertically without toppling.
    \item \textbf{Pack Shoes}: Both shoes are placed inside the shoe box and the lid is closed.
\end{itemize}

\subsection{Qualitative Results}
\label{app:qualitative-results}

\begin{figure}[htbp]
  \centering
  \includegraphics[width=\linewidth]{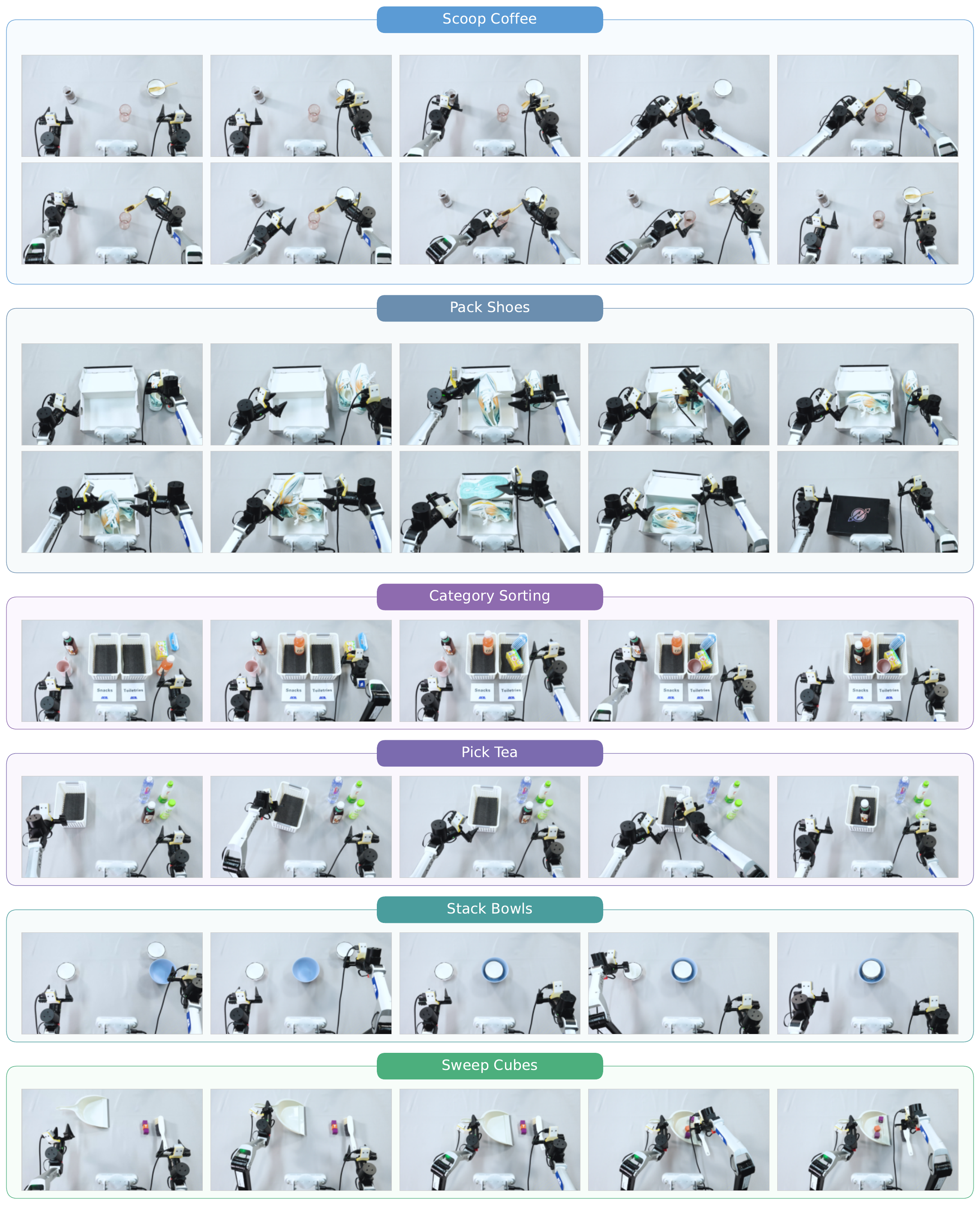}
  \caption{Qualitative rollout sequences of \Ours on the real ARX bimanual platform. Each row shows key frames from a representative task, demonstrating the policy's capability across single-arm placement, bimanual coordination, and contact-rich manipulation.}
  \label{fig:quality-results}
\end{figure}

Figure~\ref{fig:quality-results} shows qualitative rollout sequences of \Ours on the real ARX bimanual platform. Each row visualizes key frames from a successful episode, illustrating the policy's ability to execute long-horizon multi-step manipulation, bimanual coordination, and contact-rich tool use in real-world settings.

\subsection{Full RoboCasa GR1 TableTop Results}
\label{app:full-robocasa-results}

Table~\ref{tab:robocasa_full_results} reports per-task success rates on all 24 RoboCasa GR1 TableTop tasks.

\begin{table*}[tp]
\centering
\caption{Full evaluation results on the RoboCasa GR1 TableTop benchmark. Success rates (\%) over 50 rollouts per task.}
\label{tab:robocasa_full_results}
\vspace{0.3em}
\small
\renewcommand{\arraystretch}{1.1}
\setlength{\tabcolsep}{4pt}
\begin{tabular*}{\textwidth}{@{\extracolsep{\fill}}lcccccc}
\hline
\textbf{Task} & GR00T-N1.6 & Qwen3PI & FLARE & ABot-M0 & JoyAI-RA & \Ours \\
\hline
CupToDrawerClose           &  8.5 & 42.0 & 46.0 & \textbf{48.0} & \textbf{48.0} & 36.0 \\
PotatoToMicrowaveClose     & 41.5 & 42.0 & 30.0 & 50.0 & \textbf{70.0} & 58.0 \\
MilkToMicrowaveClose       & 14.0 & 50.0 & 58.0 & 46.0 & \textbf{84.0} & 56.0 \\
BottleToCabinetClose       & 51.5 & 26.0 & 66.0 & \textbf{86.0} & 84.0 & 78.0 \\
WineToCabinetClose         & 16.5 & 32.0 & 38.0 & \textbf{66.0} & 54.0 & 56.0 \\
CanToDrawerClose           & 13.0 & 62.0 & 64.0 & 74.0 & \textbf{90.0} & 70.0 \\
\hline
CuttingboardToBasket       & 58.0 & 40.0 & 44.0 & 70.0 & \textbf{88.0} & 84.0 \\
CuttingboardToCardboardbox & 46.5 & 46.0 & 54.0 & 58.0 & 46.0 & \textbf{84.0} \\
CuttingboardToPan          & 68.5 & 60.0 & 80.0 & 76.0 & \textbf{92.0} & \textbf{92.0} \\
CuttingboardToPot          & 65.0 & 40.0 & 64.0 & 66.0 & 80.0 & \textbf{84.0} \\
CuttingboardToTieredbasket & 46.5 & 44.0 & 46.0 & 38.0 & 36.0 & \textbf{72.0} \\
\hline
PlacematToBasket           & 58.5 & 44.0 & 48.0 & 52.0 & 76.0 & \textbf{86.0} \\
PlacematToBowl             & 57.5 & 52.0 & 58.0 & 66.0 & 52.0 & \textbf{72.0} \\
PlacematToPlate            & 63.0 & 50.0 & 74.0 & 60.0 & 38.0 & \textbf{80.0} \\
PlacematToTieredshelf      & 28.5 & 28.0 & 26.0 & 26.0 & 14.0 & \textbf{44.0} \\
\hline
PlateToBowl                & 57.0 & 52.0 & 50.0 & 54.0 & 48.0 & \textbf{68.0} \\
PlateToCardboardbox        & 43.5 & 40.0 & 56.0 & 48.0 & 38.0 & \textbf{70.0} \\
PlateToPan                 & 51.0 & 36.0 & \textbf{70.0} & 66.0 & 46.0 & \textbf{70.0} \\
PlateToPlate               & 78.7 & 48.0 & 76.0 & 64.0 & 88.0 & \textbf{98.0} \\
\hline
TrayToCardboardbox         & 51.5 & 34.0 & 52.0 & 54.0 & \textbf{82.0} & 78.0 \\
TrayToPlate                & 71.0 & 64.0 & 64.0 & 68.0 & 88.0 & \textbf{90.0} \\
TrayToPot                  & 64.5 & 44.0 & 70.0 & 64.0 & 88.0 & \textbf{98.0} \\
TrayToTieredbasket         & 57.0 & 50.0 & 60.0 & 60.0 & 62.0 & \textbf{74.0} \\
TrayToTieredshelf          & 31.5 & 28.0 & 28.0 & 38.0 & 24.0 & \textbf{50.0} \\
\hline
\textbf{Average}           & 47.6 & 43.9 & 55.0 & 58.3 & 63.2 & \textbf{72.8} \\
\hline
\end{tabular*}
\end{table*}

\subsection{Full RoboTwin 2.0 Results}
\label{app:full-robotwin-results}

Table~\ref{tab:robotwin2_full_task_results} reports per-task success rates on all 50 RoboTwin 2.0
tasks under both Easy/Clean and Hard/Randomized settings.

\begin{table*}[tp]
    \centering
    \caption{Full evaluation results on the RoboTwin 2.0 benchmark. Success rates are reported in
    percentage. Easy denotes the clean setting and Hard denotes the randomized setting. 100 trials per
    task.}
    \label{tab:robotwin2_full_task_results}
    \vspace{0.3em}
    \resizebox{\textwidth}{!}{
        \renewcommand{\arraystretch}{1.05}
        \begin{tabular}{l*{14}{c}}
            \toprule
        \multirow{2}{*}{Simulation Task} & \multicolumn{2}{c}{$\pi_0$} & \multicolumn{2}{c}{$\pi_{0.5}$} & \multicolumn{2}{c}{Motus} & \multicolumn{2}{c}{LingBot-VLA} & \multicolumn{2}{c}{ABot-M0} & \multicolumn{2}{c}{JoyAI-RA} & \multicolumn{2}{c}{\Ours} \\
         & Easy & Hard & Easy & Hard & Easy & Hard & Easy & Hard & Easy & Hard & Easy & Hard & Easy & Hard \\
            \midrule
        Adjust Bottle               & 99 & 95 & 100 & 99 & 89 & 93 & 100 & 100 & -- & -- & 100 & 100 & 100 & 100 \\
        Beat Block Hammer           & 79 & 84 & 96 & 93 & 95 & 88 & 92 & 89 & -- & -- & 95 & 91 & 98 & 92 \\
        Blocks Ranking RGB          & 80 & 63 & 92 & 85 & 99 & 97 & 92 & 91 & 90 & 79 & 94 & 93 & 98 & 97 \\
        Blocks Ranking Size         & 14 & 5 & 49 & 26 & 75 & 63 & 76 & 70 & -- & -- & 81 & 75 & 89 & 91 \\
        Click Alarmclock            & 77 & 68 & 98 & 89 & 100 & 100 & 97 & 43 & -- & -- & 64 & 56 & 52 & 38 \\
        Click Bell                  & 71 & 48 & 99 & 66 & 100 & 100 & 43 & 36 & -- & -- & 81 & 70 & 66 & 71 \\
        Dump Bin Bigbin             & 88 & 83 & 92 & 97 & 95 & 91 & 97 & 97 & -- & -- & 97 & 99 & 100 & 97 \\
        Grab Roller                 & 98 & 94 & 100 & 100 & 100 & 100 & 100 & 100 & -- & -- & 100 & 100 & 100 & 100 \\
        Handover Block              & 47 & 31 & 66 & 57 & 86 & 73 & 83 & 95 & 72 & 69 & 99 & 93 & 96 & 85 \\
        Handover Mic                & 97 & 97 & 98 & 97 & 78 & 63 & 94 & 99 & -- & -- & 100 & 99 & 91 & 94 \\
        Hanging Mug                 & 14 & 11 & 18 & 17 & 38 & 38 & 34 & 53 & -- & -- & 31 & 28 & 29 & 31 \\
        Lift Pot                    & 80 & 72 & 96 & 85 & 96 & 99 & 100 & 100 & -- & -- & 100 & 99 & 100 & 100 \\
        Move Can Pot                & 68 & 48 & 51 & 55 & 34 & 74 & 89 & 87 & -- & -- & 97 & 87 & 100 & 98 \\
        Move Pillbottle Pad         & 67 & 46 & 84 & 61 & 93 & 96 & 92 & 90 & 94 & 86 & 98 & 99 & 100 & 100 \\
        Move Playingcard Away       & 74 & 65 & 96 & 84 & 100 & 96 & 98 & 100 & -- & -- & 99 & 95 & 100 & 98 \\
        Move Stapler Pad            & 41 & 24 & 56 & 42 & 83 & 85 & 74 & 48 & 57 & 61 & 93 & 96 & 90 & 89 \\
        Open Laptop                 & 71 & 81 & 90 & 96 & 95 & 91 & 98 & 96 & -- & -- & 96 & 100 & 100 & 98 \\
        Open Microwave              & 4 & 32 & 34 & 77 & 95 & 91 & 91 & 92 & 88 & 84 & 97 & 99 & 91 & 85 \\
        Pick Diverse Bottles        & 69 & 31 & 81 & 71 & 90 & 91 & 88 & 85 & 71 & 65 & 85 & 90 & 84 & 86 \\
        Pick Dual Bottles           & 59 & 37 & 93 & 63 & 96 & 90 & 99 & 90 & 70 & 61 & 95 & 93 & 89 & 88 \\
        Place A2B Left              & 43 & 47 & 87 & 82 & 88 & 79 & 89 & 85 & -- & -- & 99 & 96 & 95 & 96 \\
        Place A2B Right             & 39 & 34 & 87 & 84 & 91 & 87 & 80 & 80 & -- & -- & 97 & 92 & 90 & 94 \\
        Place Bread Basket          & 62 & 46 & 77 & 64 & 91 & 94 & 95 & 93 & 89 & 86 & 88 & 91 & 92 & 93 \\
        Place Bread Skillet         & 66 & 49 & 85 & 66 & 86 & 83 & 90 & 92 & -- & -- & 92 & 89 & 94 & 89 \\
        Place Burger Fries          & 81 & 76 & 94 & 87 & 98 & 98 & 98 & 94 & -- & -- & 99 & 93 & 98 & 100 \\
        Place Can Basket            & 55 & 46 & 62 & 62 & 81 & 76 & 75 & 72 & 72 & 63 & 71 & 73 & 78 & 82 \\
        Place Cans Plasticbox       & 63 & 45 & 94 & 84 & 98 & 94 & 100 & 98 & -- & -- & 100 & 98 & 100 & 98 \\
        Place Container Plate       & 97 & 92 & 99 & 95 & 98 & 99 & 99 & 100 & -- & -- & 96 & 99 & 98 & 100 \\
        Place Dual Shoes            & 59 & 51 & 75 & 75 & 93 & 87 & 87 & 86 & 80 & 80 & 90 & 97 & 95 & 96 \\
        Place Empty Cup             & 91 & 85 & 100 & 99 & 99 & 98 & 100 & 100 & -- & -- & 100 & 100 & 100 & 100 \\
        Place Fan                   & 66 & 71 & 87 & 85 & 91 & 87 & 92 & 87 & 97 & 95 & 91 & 92 & 94 & 93 \\
        Place Mouse Pad             & 20 & 20 & 60 & 39 & 66 & 68 & 86 & 79 & -- & -- & 89 & 82 & 96 & 95 \\
        Place Object Basket         & 67 & 70 & 80 & 76 & 81 & 87 & 90 & 88 & 91 & 88 & 90 & 88 & 93 & 89 \\
        Place Object Scale          & 57 & 52 & 86 & 80 & 88 & 85 & 90 & 88 & -- & -- & 90 & 87 & 95 & 92 \\
        Place Object Stand          & 82 & 68 & 91 & 85 & 98 & 97 & 93 & 88 & 90 & 91 & 95 & 93 & 95 & 94 \\
        Place Phone Stand           & 49 & 53 & 81 & 81 & 87 & 86 & 90 & 87 & -- & -- & 95 & 95 & 91 & 98 \\
        Place Shoe                  & 76 & 76 & 92 & 93 & 99 & 97 & 99 & 99 & -- & -- & 99 & 100 & 100 & 100 \\
        Press Stapler               & 44 & 37 & 87 & 83 & 93 & 98 & 86 & 93 & -- & -- & 87 & 81 & 98 & 98 \\
        Put Bottles Dustbin         & 65 & 56 & 84 & 79 & 81 & 79 & 92 & 93 & 80 & 89 & 95 & 97 & 94 & 93 \\
        Put Object Cabinet          & 73 & 60 & 80 & 79 & 88 & 71 & 85 & 88 & -- & -- & 87 & 86 & 82 & 79 \\
        Rotate QRcode               & 74 & 70 & 89 & 87 & 89 & 73 & 86 & 82 & -- & -- & 83 & 82 & 94 & 95 \\
        Scan Object                 & 55 & 42 & 72 & 65 & 67 & 66 & 92 & 96 & 85 & 86 & 98 & 96 & 95 & 97 \\
        Shake Bottle Horizontally   & 98 & 92 & 99 & 99 & 100 & 98 & 99 & 98 & -- & -- & 100 & 100 & 100 & 100 \\
        Shake Bottle                & 94 & 91 & 99 & 97 & 100 & 97 & 100 & 99 & -- & -- & 100 & 100 & 100 & 100 \\
        Stack Blocks Three          & 72 & 52 & 91 & 76 & 91 & 95 & 96 & 95 & 84 & 77 & 60 & 62 & 87 & 82 \\
        Stack Blocks Two            & 93 & 79 & 97 & 100 & 100 & 98 & 100 & 99 & 96 & 98 & 95 & 93 & 100 & 100 \\
        Stack Bowls Three           & 77 & 75 & 77 & 71 & 79 & 87 & 71 & 77 & 80 & 86 & 80 & 81 & 80 & 85 \\
        Stack Bowls Two             & 94 & 95 & 95 & 96 & 98 & 98 & 90 & 97 & -- & -- & 95 & 93 & 96 & 98 \\
        Stamp Seal                  & 46 & 33 & 79 & 55 & 93 & 92 & 74 & 77 & 72 & 75 & 90 & 90 & 94 & 100 \\
        Turn Switch                 & 41 & 42 & 62 & 54 & 84 & 78 & 67 & 63 & 55 & 66 & 71 & 76 & 59 & 57 \\
            \midrule
        Average (\%) & 65.92 & 58.40 & 82.74 & 76.76 & 88.66 & 87.02 & 88.56 & 86.68 & 86.06 & 85.08 & 90.48 & 89.28 & \textbf{91.12} & \textbf{90.62} \\
            \bottomrule
        \end{tabular}
    }
\end{table*}

\end{document}